\documentclass{article}
\usepackage[nonatbib,final]{nips_2016}

\usepackage[utf8]{inputenc} % allow utf-8 input
\usepackage[T1]{fontenc}    % use 8-bit T1 fonts
\usepackage{hyperref}       % hyperlinks
\usepackage{url}            % simple URL typesetting
\usepackage{booktabs}       % professional-quality tables
\usepackage{amsfonts}       % blackboard math symbols
\usepackage{nicefrac}       % compact symbols for 1/2, etc.
\usepackage{microtype}      % microtypography
\usepackage{bm}
\usepackage{graphicx}
\usepackage{caption}
\usepackage{subcaption}
\newcommand{\bs}[1]{\boldsymbol{\mathbf{#1}}}
\usepackage{amsmath}
\usepackage{bbm}
\usepackage{color}
\usepackage{wrapfig}
\graphicspath{{gfx/},{gfx/supp/}}

\usepackage{tikz}

\newcommand{\secc}[1]{\autoref{sec:#1}}

\DeclareMathOperator*{\argmin}{argmin}
\newcommand{\etal}{~et al.~}

\title{Domain Separation Networks}
\author{
  Konstantinos Bousmalis\thanks{Authors contributed equally.} \\
  Google Brain\\
  Mountain View, CA \\
  \texttt{konstantinos@google.com} \\
  \And
  George Trigeorgis\footnotemark[1] \hspace{1mm}\thanks{This work was completed while George Trigeorgis was at Google Brain in Mountain View, CA.} \\
  Imperial College London \\
  London, UK\\
  \texttt{g.trigeorgis@imperial.ac.uk} \\
  \And
  Nathan Silberman \\
  Google Research \\
  New York, NY \\
  \texttt{nsilberman@google.com} \\
  \And
  Dilip Krishnan \\
  Google Research \\
  Cambridge, MA \\
   \texttt{dilipkay@google.com} \\
   \And
  Dumitru Erhan \\
  Google Brain \\
  Mountain View, CA \\
   \texttt{dumitru@google.com} \\
}
\begin{document}

\maketitle
\vspace{-4mm}
\begin{abstract}
\vspace{-4mm}
The cost of large scale data collection and annotation often makes the
application of machine learning algorithms to new tasks or datasets
prohibitively expensive. One approach circumventing this cost is training
models on synthetic data where annotations are provided automatically.
Despite their appeal, such models often fail to
generalize from synthetic to real images, necessitating
domain adaptation algorithms to manipulate these models before they can
be successfully applied. Existing approaches focus either on mapping
representations from one domain to the other, or on learning to extract features
that are invariant to the domain from which they were extracted.
However, by focusing only on creating a mapping or shared representation
between the two domains, they ignore the individual characteristics of
each domain. We suggest that explicitly modeling what is unique to each domain
can improve a model's ability to extract domain--invariant features.
Inspired by work on private--shared component analysis, we explicitly
learn to extract image representations that are partitioned into two
subspaces: one component which is private to each domain and one which
is shared across domains. Our model is trained not only to perform the
task we care about in the source domain, but
also to use the partitioned representation to reconstruct the images
from both domains. Our novel architecture results in a model that outperforms
the state--of--the--art on a range of unsupervised domain
adaptation scenarios and additionally produces visualizations of the private
and shared representations enabling interpretation of the domain adaptation
process.
\vspace{-4mm}
% to extract representations shared by both domains that are useful for
% the task of interest. We demonstrate state-of-the-art performance on several
% domain adaptation datasets.Motivated in particular by the visual unsupervised domain adaptation problem, we evaluate our method
% Recent advances in unsupervised domain adaptation have relied on strategies that attempt
% to produce representations that are agnostic to their source. This is typically
% accomplished by either mapping representations from one domain to another or by training a model to extract features such that their distributions are similar irrespective of their domain of origin. 
% Inspired by work in private--shared component analysis, we use autoencoders and subspace orthogonality constraints to explicitly model private representations for each domain that are complementary to a representation that is shared between them.  Imposing a reconstruction loss with pairs of de-correlated features allows the model
% to extract representations shared by both domains that are useful for
% the task of interest. We demonstrate state-of-the-art performance on several
% domain adaptation datasets.
\end{abstract}
\section{Introduction}
\vspace{-3mm}

The recent success of supervised learning algorithms has been partially
attributed to the large-scale datasets \cite{lin2014microsoft, ILSVRC15} 
on which they are trained. Unfortunately, collecting, annotating, and curating
such datasets is an extremely expensive and time-consuming process. An alternative
would be creating large-scale
datasets in non--realistic but inexpensive settings, such as
computer generated scenes. While such approaches offer the promise
of effectively unlimited amounts of labeled data, models trained in such
settings do not generalize well to realistic domains. 
%This is due to the fact that the distribution of low--level image statistics and high--level objects and scenes varies between domains. 
Motivated by this, we examine the problem of learning
representations that are domain–invariant in scenarios where the data
distributions during training and testing are different. In this setting,
the source data is labeled for a particular task and we would like to
transfer knowledge from the source to the target domain for which we have no ground truth labels.

%we have a labeled ---for a specific task--- dataset from a source domain, and we want to transfer knowledge from it to data from a target domain for which we have no labels. The label space is assumed to be the same between the two domains.

% Motivated by this
% shortcoming, we examine the problem of learning image representations that
% are domain--invariant. 
% More specifically, we have a dataset of images
% and ground truth annotations for a specific task (e.g. classification, segmentation)
% collected from a source domain. However, we need to apply our model using the
% same set of labels to a target domain from which we have no ground truth
% annotations to train. Consequently, our goal is to transfer our
% model's ability to perform well in the source domain to the target domain.

In this work, we focus on the tasks of object classification and pose estimation, where the object of
interest is in the foreground of a given image, for both source and target domains.
The source and target pixel distributions can differ in a number of ways. We
define ``low-level'' differences in the distributions as those arising due to noise,
resolution, illumination and color. ``High-level'' differences relate
to the number of classes, the types of objects, and geometric variations, such as
3D position and pose. %Theoretical and empirical progress in domain transfer is most tractable by selecting source and target domains that vary along specific dimensions. 
We assume that our source and target domains differ mainly in terms of the
distribution of low level image statistics and that they have high level parameters with similar distributions and the same label space. %For example, the range of poses of an object seen in the source domain is assumed to be similar to that in the target domain. We believe that this is a reasonable set of assumptions for a principled study of the visual unsupervised domain adaptation problem.

We propose a novel method, the Domain Separation Networks (DSN), for learning domain--invariant representations. Previous work attempts to either find a mapping from representations of the source domain to those of the target~\cite{sun2015return}, or find representations that are shared between the two domains~\cite{ganin2016domain,tzeng2015simultaneous,long2015learning}. While this, in principle, is a good idea, it leaves the shared representations vulnerable to contamination by noise that is correlated with the underlying shared distribution~\cite{salzmann2010factorized}. Our model, in contrast, introduces the notion of a private subspace for each domain, which captures domain specific properties, such as background and low level image statistics. A shared subspace, enforced through the use of autoencoders and explicit loss functions, captures representations shared by the domains. By finding a shared subspace that is orthogonal to the subspaces that are private, our model is able to separate the information that is unique to each domain, and in the process produce representations that are more more meaningful for the task at hand. 
Our method outperforms the state--of--the--art domain adaptation techniques on a range of datasets for object classification and pose estimation, while having an interpretability advantage by allowing the visualization of these private and shared representations. In \secc{related_work}, we survey related work and introduce relevant terminology. Our architecture, loss functions and learning regime are presented in \secc{model}. Experimental results and discussion are given in \secc{experiments}. Finally, conclusions and directions for future work are in \secc{conclusion}.
\vspace{-4mm}
\section{Related Work}
\vspace{-4mm}
\label{sec:related_work}

Learning to perform unsupervised domain adaptation is an open theoretical and practical problem. While much prior art exists, our literature 
review focuses primarily on Convolutional Neural Network (CNN) based methods due
to their empirical superiority on this problem~\cite{ganin2016domain, long2015learning, sun2015return, tzeng2015ddc}. Ben-David \etal \cite{ben2010theory} provide upper bounds on a domain-adapted
classifier in the target domain. They introduce the idea of training a binary 
classifier trained to distinguish source and target domains. The error that this
``domain incoherence'' classifier provides (along with the error of a source domain specific classifier) combine to give the overall bounds. Mansour \etal \cite{mansour2009domain} extend the theory of \cite{ben2010theory} to handle the case of multiple source domains. 

Ganin \etal ~\cite{ganin2014unsupervised,ganin2016domain} and Ajakan
\etal ~\cite{Ajakan2014} use adversarial training to find
domain--invariant representations
in-network. Their Domain--Adversarial Neural Networks (DANN) exhibit an 
architecture whose first few feature extraction layers are shared by two
classifiers trained simultaneously. The first is trained to correctly
predict task-specific class labels on the source data  while the second is
trained to predict the domain of each input. DANN minimizes the domain
classification loss with respect to parameters specific to the domain
classifier, while maximizing it with respect to the parameters that are
common to both classifiers. This minimax optimization becomes possible via
the use of a  gradient reversal layer (GRL). 

%While their model demonstrated the
%mechanism by which a domain-incoherence loss can be used in-network, it does not distinguish between elements that the model
%should be able to distinguish (domain-private aspects of the representation)
%and elements that the model should not be able to distinguish (domain-common
%aspects of the representation).

% Let $\hat{d} \in \{S, T\}$ be the categorical random variable that denotes the domain prediction for each sample. The Domain--Adversarial Network \cite{ganin2014unsupervised} tries to match the conditional distributions $p_S(\hat{d}\;|\;{\bf X}_S; {\bf W}_f, {\bf W}_d)$ and $p_T(\hat{d}\;|\;{\bf X}_T; {\bf W}_f, {\bf W}_d)$, where ${\bf W}_f$ are the CNN weights for the domain--invariant feature extraction layers, ${\bf W}_d$ are the weights of the domain classifier. Note that ${\bf W}_f$ and ${\bf W}_d$ are shared for both domains.

Tzeng \etal \cite{tzeng2015ddc} and Long \etal
\cite{long2015learning} proposed versions of this model where
the maximization of the domain classification loss is replaced by the minimization
of the Maximum Mean Discrepancy (MMD) metric \cite{gretton2012mmd}. The MMD
metric is computed between features extracted from sets of samples from each
domain. The Deep Domain Confusion Network by Tzeng \etal \cite{tzeng2015ddc} has an 
MMD loss at one layer in the CNN architecture while Long \etal
\cite{long2015learning} proposed the Deep Adaptation Network that has MMD losses at
multiple layers. 

% NS: removed as this criticism applies partially to our method as well and
% this does not serve as a differentiator.
%Both the domain--adversarial and the MMD--based networks are somewhat more difficult to train than regular CNNs. The former involve a minimax optimization and the latter requires finding an optimal kernel that maximizes the power of the two-sample test, which is not trivial as this is not guaranteed to remain optimal throughout training. 

Other related techniques involve learning a transformation from one domain to the other. In this setup, the feature extraction pipeline is fixed during the domain adaptation optimization. 
This has been applied in various non-CNN based approaches
\cite{gong2012geodesic,caseiro2015beyond,gopalan2011domain} as well as the recent
CNN-based Correlation Alignment (CORAL) \cite{sun2015return} algorithm which
``recolors'' whitened source features with the covariance of features from the
target domain.

\section{Method}
\vspace{-4mm}
\label{sec:model}
\begin{figure}[tb]
     \centering
     \includegraphics[width=1\linewidth]{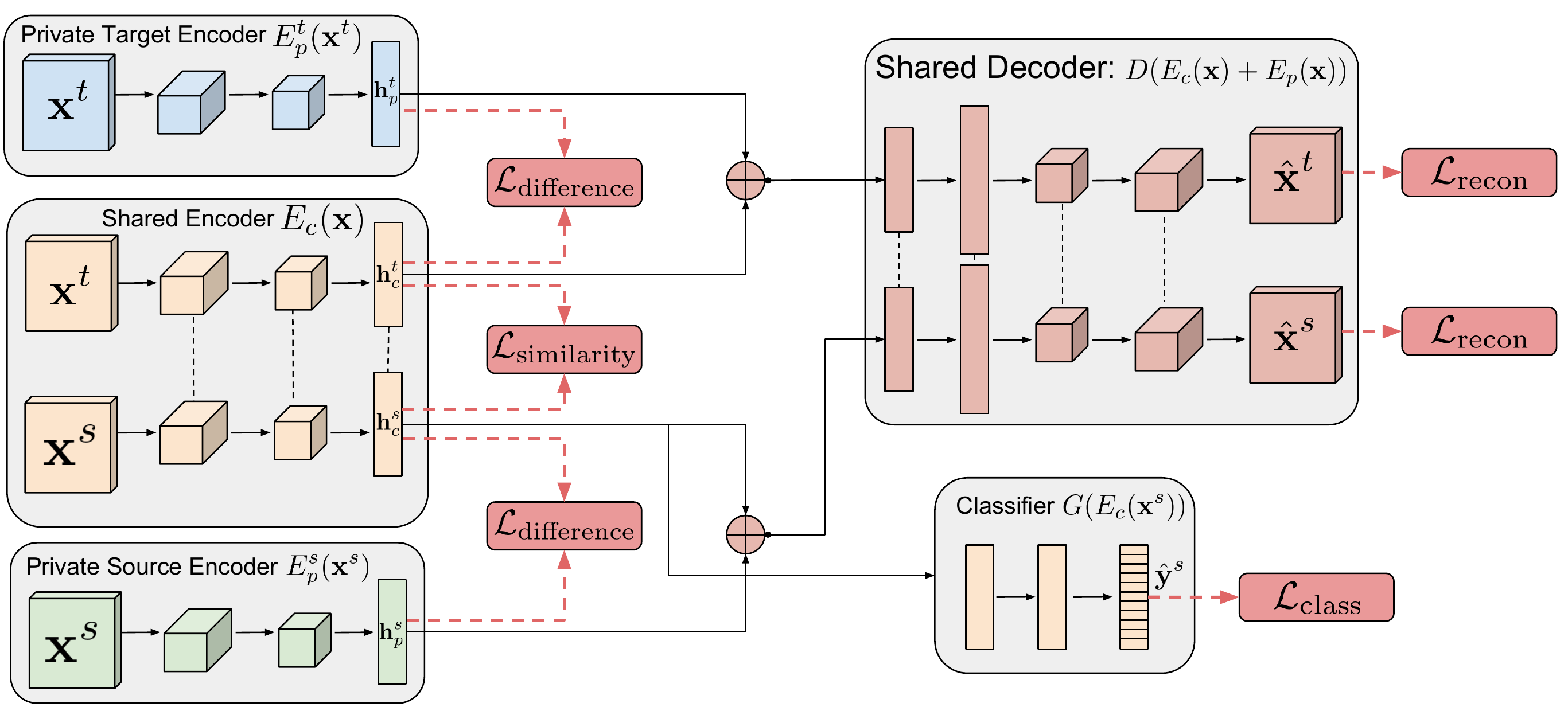}
     \caption{Training of our Domain Separation Networks. A shared-weight encoder $E_c(\bs x)$ learns to capture representation components for a given input sample that are shared among domains. A private encoder $E_p(\bs x)$ (one for each domain) learns to capture domain--specific components of the representation. A shared decoder learns to reconstruct the input sample by using both the private and source representations. The private and shared representation components are pushed apart with soft subspace orthogonality constraints $\mathcal {L}_\mathrm{difference}$, whereas the shared representation components are kept similar with a similarity loss $\mathcal {L}_\mathrm{similarity}$. See text for more information.}
     \label{fig:model}
\end{figure}
While the Domain Separation Networks (DSNs) could in principle be applicable to other learning tasks, without loss of generalization, we mainly use
image classification as the cross-domain task. Given a labeled dataset in a source domain and
an unlabeled dataset in a target domain, our goal is to train a classifier on data
from the source domain that generalizes to the target domain. Like previous efforts~\cite{ganin2014unsupervised,ganin2016domain},
our model is trained such that the representations of images from the source
domain are similar to those from the target domain. This allows
a classifier trained on images from the source domain to generalize
as the inputs to the classifier are in theory invariant to the domain of origin. However, these representations might trivially include noise that is highly correlated with the shared
representation, as shown by Salzmann \etal ~\cite{salzmann2010factorized}.

Our main novelty is that, inspired by recent
work \cite{jia2010factorized,salzmann2010factorized,virtanen2011bayesian} on 
shared--space component analysis, DSNs explicitly and jointly model both private and shared components of the domain representations. The private component
of the representation is specific to a single domain and the shared component
of the representation is shared by both domains. To induce the model to 
produce such split representations, we add a loss function that encourages
independence of these parts. Finally, to ensure that the private
representations are still useful (avoiding trivial solutions) and to add
generalizability, we also add a reconstruction loss.
The combination of these objectives is a model that produces a shared
representation that is similar for both domains and a private representation that
is different. By partitioning the space in such a manner, the classifier trained on
the shared representation is better able to generalize across domains as its inputs
are uncontaminated with aspects of the representation that are unique to each domain.

% Our task is classification in a target domain for which we have no labeled data. Our goal is to train a feature extractor that provides representations that are similar for related samples from the source and target domains. Previous work in unsupervised domain adaptation, as discussed in Section \secc{related_work}, attempts to find representations of the input samples that are domain--invariant. In this work, motivated by ideas from private--shared space models \cite{jia2010factorized,salzmann2010factorized,virtanen2011bayesian}, we suggest separating these representations to components that are shared between domains and to components that are private to each domain. By doing so, we are able to produce shared components that are more domain--invariant. In our model, as one can see in Fig.~\ref{fig:model}, we explicitly and jointly model both  private and shared representations of the domains. 

More specifically, 
let ${\bs X}_S = \{({\bs x}_i^s, {\bs y}_i^s)\}_{i=0}^{N_s}$ represent a labeled dataset
of $N_s$ samples from the source domain where ${\bs x}_i^s \sim {\cal D}_S$ and let
${\bs X}^t = \{{\bs x}_i^t\}_{i=0}^{N_t}$ represent an unlabeled dataset of $N_t$
samples from the target domain where ${\bs x}_i^t \sim {\cal D}_T$. Let
$E_c(\bs x;\bs \theta_c)$ be a function parameterized by $\bs \theta_c$ which maps
an image $\bs x $ to a hidden representation $\bs h_c$ representing features 
that are common or
\textit{shared} across domains. Let $E_p(\bs x;\bs \theta_p)$ be
an analogous function which maps an image $\bs x$ to a hidden representation
$\bs h_p$ representing features that are \textit{private} to each domain.
Let $D(\bs h;\bs  \theta_d)$ be a decoding function mapping a hidden representation
$\bs h$ to an image reconstruction $\hat{\bs x}$. Finally, $G(\bs h; \bs \theta_g)$
represents a task-specific function, parameterized by $\bs \theta_g$ that maps from
hidden representations $\bs h$ to the task-specific predictions $\hat{\bs y}$. The resulting Domain Separation Network (DSN) model is depicted in \autoref{fig:model}.

\subsection{Learning}
\label{sec:learning}
Inference in a DSN model is given by $\hat{\bs x} = D(E_c(\bs x) + E_p(\bs x))$
and $\hat{\bs y} = G(E_c(\bs x))$
where $\hat{\bs x}$ is the reconstruction of the input $\bs x$ and $\hat{\bs y}$ is the
task-specific prediction. 
% Nathan's version 
% \subsection{Learning}
% The goal of training is to minimize the following loss:
% \begin{align}
% \mathcal{L} \;=\; & \mathcal {L}_\mathrm{class}(\bs y^s, \hat{\bs y}^s) +
% \alpha \;(\mathcal {L}_\mathrm{recon}(\bs x^s, \hat{\bs x}^s) \;+ \; \mathcal {L}_\mathrm{recon}(\bs x^t, \hat{\bs x}^t))\; + \\ \nonumber
% & \beta\; \mathcal ({L}_\mathrm{difference}(\bs h_c^s, \bs h_p^s) \; + \; \mathcal {L}_\mathrm{difference}(\bs h_c^t, \bs h_p^t))\; +\;
% \gamma\; \mathcal {L}_\mathrm{similarity}(\bs h_c^s, \bs h_c^t)
% \end{align}
% The \textbf{classification} loss $\mathcal{L}_\mathrm{class}$ is applied to the
% source domain only and utilizes the popular cross entropy loss. The
% \textbf{reconstruction} loss
% $\mathcal{L}_\mathrm{recon}(\bs x, \hat{\bs x})=\| \bs x\;-\;\hat{\bs x}\|_2^2$ is applied to both domains.
% The \textbf{difference} loss encourages the shared and private encoders to encode different
% elements of the inputs and is defined in Section \ref{sec:difference_loss}. Finally, the
% \textbf{similarity} loss encourages the hidden representations from the shared encoder
% to be as 
% similar as possible irrespective of the domain. It is defined in
% Section \ref{sec:similarity_loss} 
The goal of training is to minimize the following loss with respect to parameters ${\bs \Theta = \{\bs \theta_c, \bs \theta_p, \bs \theta_d, \bs \theta_g\}}$:
\begin{align}
\mathcal{L} \;=\; & \mathcal {L}_\mathrm{task} +
\alpha \;\mathcal {L}_\mathrm{recon}\;+ \;  \beta\; \mathcal {L}_\mathrm{difference}\; +\;
\gamma\; \mathcal {L}_\mathrm{similarity}
\end{align}
 where $\alpha, \beta, \gamma$ are weights that control the interaction of the loss terms.
%%%%%%%%%%%%%%%%%%%%%%%
% CLASSIFICATION LOSS %
%%%%%%%%%%%%%%%%%%%%%%%
 The classification loss $\mathcal{L}_\mathrm{task}$ trains the model
to predict the output labels we are ultimately interested in. Because we assume
the target domain is unlabeled, the loss is applied only to the source domain.
We want to minimize the negative log--likelihood of the ground truth class for each source domain sample:
\begin{equation}
\mathcal {L}_\mathrm{task} = -\sum_{i=0}^{N_s}\bs y_i^s\cdot \log \hat{\bs y}_i^s,
\end{equation}
where $\bs{y}_i^s$ is the one--hot encoding of the class label for source input $i$
and $\hat{\bs y}_i^s$ are the softmax predictions of the model: $\hat{\bs y}_i^s = G(E_c(\bs x_i^s))$.
%%%%%%%%%%%%%%%%%%%%%%%
% RECONSTRUCTION LOSS %
%%%%%%%%%%%%%%%%%%%%%%%
We use a scale--invariant mean squared error term~\cite{eigen2014depth} for the
reconstruction loss $\mathcal{L}_\mathrm{recon}$ which is applied to
both domains:
\begin{align}
\label{eq:reconstruction_loss}
\mathcal{L}_\mathrm{recon} &=
\sum_{i=1}^{N_s} \mathcal{L}_\mathrm{si\_mse}({\bs x}_i^s, \hat{{\bs x}}_i^s)
+ \sum_{i=1}^{N_t} \mathcal{L}_\mathrm{si\_mse}({\bs x}_i^t, \hat{{\bs x}}_i^t) \\
\mathcal{L}_\mathrm{si\_mse}({\bs x}, \hat{{\bs x}}) &=
\frac{1}{k} \|{\bs x}-\hat{{\bs x}}\|_2^2 - \frac{1}{k^2}([{\bs x}-\hat{{\bs x}}] \cdot {\bs 1}_k)^2,
\end{align}
where  $k$ is the number of pixels in input $x$, $\bs 1_k$ is a vector of ones of
length $k$; and $\|\cdot\|_2^2$ is the squared $L_2$-norm. While a mean squared error
loss is traditionally used for reconstruction tasks, it penalizes predictions that
are correct up to a scaling term. %A prediction $\hat{\bs{x}}=\bs{x}+5$ would incur a loss of $25k$ where $k$ is the number of pixels in $\bs{x}$. 
Conversely, the
scale-invariant mean squared error penalizes differences between \textit{pairs}
of pixels. This allows the model to learn to reproduce the overall shape of the objects
being modeled without expending modeling power on the absolute color or intensity
of the inputs. We validated that this reconstruction loss was indeed the correct choice experimentally in Section~\ref{sec:discussion} by training a version of our best DSN model with the traditional mean squared error loss instead of the one in \autoref{eq:reconstruction_loss}.

%%%%%%%%%%%%%%%%%%%
% DIFFERENCE LOSS %
%%%%%%%%%%%%%%%%%%%
The difference loss is also applied to both domains and encourages the
shared and private encoders to encode different aspects of the inputs. We define
the loss via a soft subspace orthogonality constraint between the
private and shared representation of each domain. Let $\bs H_c^s$ and $\bs H_c^t$ 
be matrices whose rows are the hidden \textit{shared} representations
$\bs h_c^s = E_c({\bs x}^s)$ and $\bs h_c^t = E_c({\bs x}^t)$ from samples of source
and target data respectively. Similarly, let $\bs H_p^s$ and $\bs H_p^t$ be matrices
whose rows are the \textsl{private} representation $\bs h_p^s=E_p^s({\bs x}^s)$ and
$\bs h_p^t=E_p^t({\bs x}^t)$ from samples of source and target data respectively. The
difference loss encourages orthogonality between the shared and the private
representations of each domain:
\begin{equation}
L_\mathrm{difference}\;=\; \left\|{\bs H_c^s}^\top \bs H_p^s\right\|_F^2 \; + \; \left\|{\bs H_c^t}^\top \bs H_p^t\right\|_F^2,
\end{equation}
where $\|\cdot\|_F^2$ is the squared Frobenius norm.
Finally, the similarity loss encourages the hidden
representations ${\bs h}_c^s$ and ${\bs h}_c^t$ from the shared encoder to be as
similar as possible irrespective of the domain. We experimented with two similarity losses, which we discuss in detail.

%%%%%%%%%%%%%%%%%%%
% SIMILARITY LOSS %
%%%%%%%%%%%%%%%%%%%

\subsection{Similarity Losses}
\label{sec:similarity_loss}
The domain adversarial similarity loss \cite{ganin2014unsupervised,ganin2016domain}
is used to train a model to produce representations such that a classifier cannot
reliably predict the domain of the encoded representation.
Maximizing such ``confusion'' is achieved via a Gradient Reversal Layer (GRL)
and a \textit{domain classifier} trained to predict the domain producing the
hidden representation.
The GRL has the same output as the identity function,
but reverses the gradient direction. Formally, for some function $f(\bs u)$, the GRL
is defined as $Q\left(f(\bs u)\right) = f(\bs u)$ with a gradient ${\frac{d}{d\bs u}Q(f(\bs u))=-\frac{d}{d\bs u}f(\bs u)}$.
% \begin{align}
%     Q\left(f(\bs u)\right) = f(\bs u), \hspace{20pt}
%     \frac{d}{d\bs u}Q(f(\bs u))=-\frac{d}{d\bs u}f(\bs u)
% \end{align}
The domain classifier $Z(Q(\bs h_c); \bs \theta_z) \rightarrow \hat{d}$
parameterized by $\bs \theta_z$ maps a shared representation vector
$\bs h_c=E_c(\bs x ; \bs \theta_c)$ to a prediction of the label $\hat{d} \in \{0,1\}$ of the
input sample ${\bs x}$. Learning with a GRL is adversarial in that $\bs \theta_z$ is
optimized to increase $Z$'s ability to discriminate between encodings of images from
the source or target domains, while the reversal of the gradient results in the model
parameters $\bs \theta_c$ learning representations from which domain classification
accuracy is reduced.
Essentially, we \textsl{maximize} the binomial cross--entropy for the domain prediction task with respect to $\bs \theta_z$, while \textsl{minimizing} it with respect to $\bs \theta_c$: 
\begin{equation}
\label{eq:dann}
\mathcal {L}_\mathrm{similarity}^\mathrm{DANN} = \;\sum_{i=0}^{N_s+N_t} \left\{d_i \log \hat{d}_i \;+\;  (1-d_i) \log(1- \hat{d}_i)\right \}.
\end{equation}

where $d_i \in \{0,1\}$ is the ground truth domain label for sample $i$.

The Maximum Mean Discrepancy (MMD) loss \cite{gretton2012mmd} is a kernel-based
distance function between pairs of samples.
We use a biased statistic for the squared population MMD between shared encodings of the source samples $\bs h_c^s$ and the shared encodings of the target samples $\bs h_c^t$:
\begin{equation}
\label{eq:mmd}
\mathcal {L}_\mathrm{similarity}^\mathrm{MMD}\;=\; \displaystyle \frac{1}{(N^s)^2} \sum_{i,j=0}^{N^s}\kappa(\bs h_{ci}^s, \bs h_{cj}^s)-\frac{2}{N^sN^t}\sum_{i,j=0}^{N^s,N^t}\kappa(\bs h_{ci}^s, \bs h_{cj}^t) + \frac{1}{(N^t)^2}\sum_{i,j=0}^{N^t}\kappa(\bs h_{ci}^t,\bs h_{cj}^t),
\end{equation}
where $\kappa(\cdot,\cdot)$ is a PSD kernel function. In our experiments we used a linear combination of multiple RBF kernels: $\kappa(x_i, x_j) = \sum_n \eta_n \exp \{-\frac{1}{2\sigma_n}\|\bs x_i - \bs x_j\|^2\}$, where $\sigma_n$ is the standard deviation and $\eta_n$ is the weight for our $n^{th}$ RBF kernel. Any additional kernels we include in the multi--RBF kernel are additive and guarantee that their linear combination remains characteristic. Therefore, having a large range of kernels is beneficial since the distributions of the shared features change during learning, and different components of the multi--RBF kernel might be responsible at different times for making sure we reject a false null hypothesis, i.e. that the loss is sufficiently high when the distributions are not similar~\cite{long2015learning}. The advantage of using an RBF kernel with the MMD distance is that the Taylor expansion of the Gaussian function allows us to match all the moments of the two populations. The caveat is that it requires finding optimal kernel bandwidths $\sigma_n$.

\section{Evaluation}
\label{sec:experiments}
\vspace{-4mm}

% We evaluate our model on several domain adaptation scenarios
% namely MNIST and MNIST-M \cite{ganin2016domain}, Synthetic to Real German Traffic Signs from the \textsl{German Traffic Signs
% Recognition Benchmark} (GTSRB) \cite{stallkamp2012gtsrb} and Streetview
% House Numbers (SVHN) \cite{netzer2011svhn}. We also used the LineMod dataset
% \cite{hinterstoisser2012accv} it serves as a good proof-of-concept for the
% problem sphere we are ultimately interested in, namely training on synthetic
% and testing on real images.{\color{green} TODO(nsilberman): Make sure this message comes across early, possibly
% even in the abstract} {\color{blue} Agreed!}

% We are compelled to mention that while the Office
% our dataset based on the LineMod dataset \cite{hinterstoisser2012accv}. 

We are motivated by the problem of learning models on a clean, synthetic dataset and testing on
noisy, real–world dataset. To this end, we evaluate on object classification datasets used in previous work\footnote{The most commonly used dataset for visual domain adaptation in the context of object classification is Office \cite{saenko2010adapting}. However, this dataset exhibits significant variations in both low-level and high-level parameter distributions. Low-level variations are due to the different cameras and background textures in the images (e.g. Amazon versus DSLR). However, there are significant high-level variations due to object identity: e.g. the motorcycle class contains non-motorcycle objects; the backpack class contains a laptop; some domains contain the object in only one pose. Other commonly used datasets such as Caltech-256 suffer from similar problems. We therefore exclude these datasets from our evaluation. For more information, see our Supplementary Material.}
including MNIST and {MNIST-M} \cite{ganin2016domain}, the German Traffic Signs
Recognition Benchmark (GTSRB) \cite{stallkamp2012gtsrb}, and the Streetview
House Numbers (SVHN) \cite{netzer2011reading}. We also evaluate on the cropped
LINEMOD dataset, a standard for object instance
recognition and 3D pose 
estimation~\cite{hinterstoisser2012accv,wohlhart2015learning}, for which we have
synthetic and real data\footnote{\url{https://cvarlab.icg.tugraz.at/projects/3d_object_detection/}}.
We tested the following unsupervised domain adaptation scenarios: \textsl{(a)} from MNIST to MNIST-M; \textsl{(b)} from SVHN to MNIST; \textsl{(c)} from synthetic traffic signs to real ones with GTSRB; \textsl{(d)} from synthetic LINEMOD object instances rendered on a black background to the same object instances in the real world.

We evaluate the efficacy of our method with each of the two similarity losses outlined in \autoref{sec:similarity_loss} by comparing against the prevailing visual domain adaptation techniques for neural networks: Correlation Alignment (CORAL)~\cite{sun2015return},  Domain--Adversarial Neural Networks (DANN)~\cite{ganin2014unsupervised,ganin2016domain}, and MMD regularization~\cite{tzeng2015ddc,long2015learning}. 
For each scenario we provide two additional baselines: the performance on the target domain of the respective model with no domain adaptation and trained \textsl{(a)} on the source domain (``Source--only'' in \autoref{tab:results}) and \textsl{(b)} on the target domain (``Target--only''), as an empirical lower and upper bound respectively. 

We have not found a universally applicable way to optimize hyperparameters for unsupervised domain adaptation.  Previous work~\cite{ganin2016domain} suggests the use of reverse validation. We implemented this (see Supplementary Material for details) but found that that the reverse validation accuracy did not always align well with test accuracy. Ideally we would like to avoid using labels from the target domain, as it can be argued that if ones does have target domain labels, they should be used during training. However, there are applications where a labeled target domain set cannot be used for training. An example is the labeling of a dataset with the use of AprilTags~\cite{olson2011apriltag}, 2D barcodes that can be used to label the pose of an object, provided that a camera is calibrated and the physical dimensions of the barcode are known. These images should not be used when learning features from pixels, because the model might be able to decipher the tags. However, they can be part of a test set that is not available during training, and an equivalent dataset without the tags could be used for unsupervised domain adaptation.
 We thus chose to use a small set of labeled target domain data as a validation set for the hyperparameters of all the methods we compare. All methods were evaluated using the same protocol, so comparison numbers are fair and meaningful. The performance on this validation set can serve as an \textsl{upper bound} of a satisfactory validation metric for unsupervised domain adaptation, which to our knowledge is still an open research question, and out of the scope of this work.

\begin{table}[t]
\centering
\caption{Mean classification accuracy (\%) for the unsupervised domain adaptation scenarios we evaluated all the methods on. We have replicated the experiments from Ganin \etal~\protect\cite{ganin2016domain} and in parentheses we show the results reported in their paper. The ``Source--only'' and ``Target--only'' rows are the results on the target domain when  using no domain adaptation and training only on the source or the target domain respectively. The results that perform best in each domain adaptation task are in bold font.}
\vspace{2mm}
\label{tab:results}
\begin{tabular}{ | l | l | l | l | l | }
\hline
\bf Model   & \bf MNIST to & \bf Synth Digits to &\bf SVHN to  &\bf Synth Signs to \\
 &\bf MNIST-M  &\bf SVHN        &\bf MNIST &\bf GTSRB     \\ \hline \hline
Source-only  & 56.6 (52.2) & 86.7 (86.7)      & 59.2 (54.9) & 85.1  (79.0)    \\ \hline \hline
CORAL \cite{sun2015return} & 57.7 & 85.2       &  63.1     & 86.9        \\ \hline
MMD  \cite{tzeng2015ddc,long2015learning}  & 76.9 & 88.0 & 71.1 & 91.1 \\ \hline 
DANN \cite{ganin2016domain}  & 77.4 (76.6)  &  90.3 (91.0)    & 70.7 (73.8)  &     92.9 (88.6) \\ \hline
DSN w/ MMD (ours)  & 80.5 & 88.5  & 72.2   & 92.6 \\ \hline
DSN w/ DANN (ours) & \textbf{83.2} & \textbf{91.2} & \textbf{82.7} & \textbf{93.1} \\ \hline\hline
Target-only  & 98.7 & 92.4  & 99.5 & 99.8  \\ \hline
\end{tabular}
\end{table}
%We reproduce the MNIST-M dataset from the instructions in~\cite{ganin2016domain}. 
 % Ganin uses 32x32 images and 59001 from training and 9001 for evaluation.

\subsection{Datasets and Adaptation Scenarios}
\textbf{MNIST to MNIST-M.} In this domain adaptation scenario we use the popular MNIST~\cite{lecun1998gradient} dataset of handwritten digits as the source domain, and MNIST-M, a variation of MNIST proposed for unsupervised domain adaptation by ~\cite{ganin2016domain}. MNIST-M was created by using each MNIST digit as a binary mask and inverting with it the colors of a background image. The background images are random crops uniformly sampled from the Berkeley Segmentation Data Set~(BSDS500)~\cite{arbelaez2011contour}.  In all our experiments, following the experimental protocol by~\cite{ganin2016domain}. Out of the $59,001$ MNIST-M training examples, we used the labels for $1,000$ of them to find optimal hyperparameters for our models. This scenario, like all three digit adaptation scenarios, has 10 class labels.
%We reproduce the MNIST-M dataset from the instructions in~\cite{ganin2016domain}. 
 % Ganin uses 32x32 images and 59001 from training and 9001 for evaluation.

\textbf{Synthetic Digits to SVHN.}  In this scenario we aim to learn a classifier for the Street-View House Number data set (SVHN)~\cite{netzer2011reading}, our target domain, from a dataset of purely synthesized digits, our source domain. The synthetic digits~\cite{ganin2016domain} dataset was created by rasterizing bitmap fonts in a sequence~(one, two, and three digits) with the ground truth label being the digit in the center of the image, just like in SVHN. The source domain samples are further augmented by variations in scale, translation, background colors, stroke colors, and Gaussian blurring. We use $479,400$ Synthetic Digits for our source domain training set, $73,257$ unlabeled SVHN samples for domain adaptation, and $26,032$ SVHN samples for testing.  Similarly to above, we used the labels of $1,000$ SVHN training examples to find optimal hyperparameters for our models. 

\textbf{SVHN to MNIST.}  Although the SVHN dataset contains significant variations (in scale, background clutter, blurring, embossing, slanting, contrast, rotation, sequences to name a few) there is not a lot of variation in the actual digits shapes. This makes it quite distinct from a dataset of handwritten digits, like MNIST, where there are a lot of elastic distortions in the shapes, variations in thickness, and noise on the digits themselves. Since the ground truth digits in both datasets are centered, this is a well--posed and rather difficult domain adaptation scenario. As above, we used the labels of $1,000$ MNIST training examples for validation.

\textbf{Synthetic Signs to GTSRB.} We also perform an experiment using a dataset of synthetic traffic signs from \cite{Moiseev2013} to real world dataset of traffic signs (GTSRB)~\cite{stallkamp2012gtsrb}. While the three digit adaptation scenarios have 10 class labels, this scenario has 43 different traffic signs. The synthetic signs were obtained by taking relevant pictograms and adding various types of variations, including random backgrounds, brightness, saturation, 3D rotations, Gaussian and motion blur. We use $90,000$ synthetic signs for training, $1,280$ random GTSRB real--world signs for domain adaptation and validation, and the remaining $37,929$ GTSRB real signs as the test set.

\textbf{Synthetic Objects to LineMod.} The LineMod dataset \cite{wohlhart2015learning} consists of CAD models of objects in a cluttered environment and a high variance of 3D poses for each object. We use the 11 non--symmetric objects from the cropped version of the dataset, where the images are cropped with the object in the center, for the task of object instance recognition and 3D pose estimation. We train our models on $16,962$ images for these objects rendered on a black background without additional noise. We use a target domain training set of $10,673$ real--world images for domain adaptation and validation, and a target domain test set of $2,655$ for testing. For this scenario our task is both classification and pose estimation; our task loss is therefore $\mathcal{L}_\mathrm{task}=\sum_{i=0}^{N_s}\{-\bs y_i^s\cdot \log \hat{\bs y}_i^s+\xi \log(1-|\bs q^s\cdot  \hat{\bs q}^s|)\}$,
where $\bs{q}^s$ is the positive unit quaternion vector representing the ground truth 3D pose, and $\hat{\bs q}^s$ is the equivalent prediction. The first term is the classification loss, similar to the rest of the experiments, the second term is the log of a 3D rotation metric for quaternions~\cite{huynh2009metrics}, and $\xi$ is the weight for the pose loss. Quaternions are a convenient angle--axis representation for 3D rotations. In \autoref{tab:pose_results} we report the mean angle the object would need to be rotated (on a fixed 3D axis) to move from the predicted pose to the ground truth \cite{hinterstoisser2012accv}.

%%% Original Table submitted to NIPS
% \begin{table}[]
% \centering
% \caption{Mean classification accuracy and pose error for the ``Synth Objects to LINEMOD'' scenario. }
% \label{tab:pose_results}
% \begin{tabular}{@{}|l|c|c|@{}}
% \hline
% \multicolumn{1}{|c|}{\textbf{Method}} & \multicolumn{1}{c|}{\textbf{Classification Accuracy \%}} & \multicolumn{1}{c|}{\textbf{Mean Angle Error in Radians}} \\ \hline \hline
% Source-only                           & 47.33                                   &1.66\\\hline  \hline%         & 89.2                                     \\\hline\hline
% MMD                                   & 72.35                                        &1.23\\\hline%    & 70.62                              \\ \hline
% DANN                                  & 99.90                                               &1.09\\\hline%& 62.72                  \\ \hline\hline
% DSN w/ MMD (ours)                     & 99.72                                            &1.16\\\hline%& 66.49                                    \\ \hline
% DSN w/ DANN (ours)                    & 99.97                                           &1.086\\\hline  \hline% & 62.27                                     \\ \hline\hline
% Target-only                           & 100.00                                              &0.11\\\hline%& 6.47                                  \\ \hline
% \end{tabular}
% \end{table}

\begin{table}[]
\centering
\caption{Mean classification accuracy and pose error for the ``Synth Objects to LINEMOD'' scenario. }
\label{tab:pose_results}
\begin{tabular}{@{}|l|c|c|@{}}
\hline
\multicolumn{1}{|c|}{\textbf{Method}} & \multicolumn{1}{c|}{\textbf{Classification Accuracy}} & \multicolumn{1}{c|}{\textbf{Mean Angle Error}} \\ \hline \hline
Source-only                           & 47.33\%                                   & $89.2^{\circ}$\\\hline  \hline
MMD                                   & 72.35\%                                        & $70.62^{\circ}$                              \\ \hline
DANN                                  & 99.90\%                                               &$56.58^{\circ}$                 \\ \hline\hline
DSN w/ MMD (ours)                     & 99.72\%                                            &$66.49^{\circ}$\\ \hline
DSN w/ DANN (ours)                    & \textbf{100.00}\%                                           & $\textbf{53.27}^{\circ}$\\ \hline\hline
Target-only                           & 100.00\%                                              &$6.47^{\circ}$                                  \\ \hline
\end{tabular}
\end{table}
\subsection{Implementation Details}
\begin{figure}[t]
    \centering
    \begin{subfigure}[b]{.23\linewidth}
        \centering
        \includegraphics[height=\linewidth]{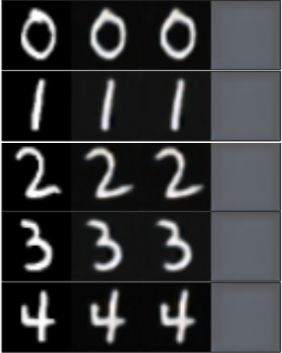}
        \caption{MNIST (source)}
        \label{fig:mnist_to_mnist-m_source}
    \end{subfigure}          \hfill
    \begin{subfigure}[b]{.23\linewidth}
        \centering
    \includegraphics[height=\linewidth]{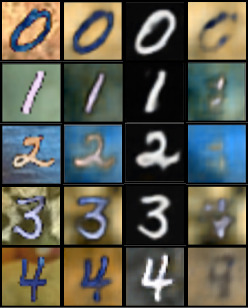}
        \caption{MNIST-M (target)}
        \label{fig:mnist_to_mnist-m_target}
    \end{subfigure}          \hfill
    \begin{subfigure}[b]{.23\linewidth}
        \centering
        \includegraphics[height=\linewidth]{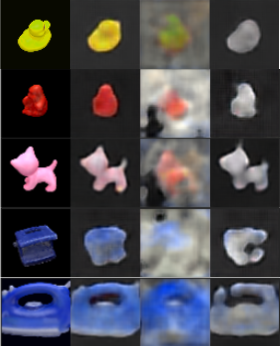}
        \caption{Synth Objects (source)}
        \label{fig:mnist_to_mnist-m_target}
    \end{subfigure}          \hfill
    \begin{subfigure}[b]{.23\linewidth}
        \centering
        \includegraphics[height=\linewidth]{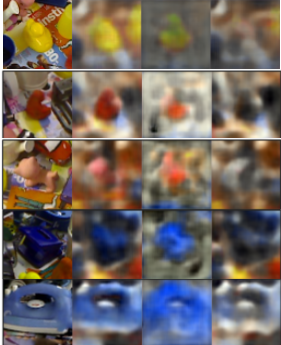}
        \caption{LINEMOD (target)}
        \label{fig:mnist_to_mnist-m_target}
    \end{subfigure}          \hfill    
    \caption{Reconstructions for the representations of the two domains for ``MNIST to MNIST-M'' and for ``Synth Objects to LINEMOD''. In each block from left to right: the original image $\bs x_t$; reconstructed image $D({E_c(\bs x^t) + E_p(\bs x^t)}) $; shared only reconstruction  $D(E_c(\bs x^t)) $;  private only reconstruction  $D(E_p(\bs x^t))$.}
    \label{fig:mnist_m_visualizations}
\end{figure}
All the models were implemented using TensorFlow\footnote{Our code will be open--sourced under \url{https://github.com/tensorflow/models/} before the NIPS 2016 meeting.}~\cite{abadi2016tensorflow} and were trained with Stochastic Gradient Descent plus momentum~\cite{sutskever2013momentum}. Our initial learning rate was multiplied by $0.9$ every $20,000$ steps (mini-batches). We used batches of 32 samples from each domain for a total of 64 and
the input images were mean-centered and rescaled to $[-1, 1]$.
In order to avoid distractions for the main classification task during the early stages of the training procedure, we activate any additional domain adaptation loss after $10,000$ steps of training. 
%Preliminary experiments with gradually changing the additional losses coefficients from $0$ to an optimal value, validated experimentally, did not show any performance improvement, so we instead opted for simplicity. Activating the additional losses even later did not have a performance effect for our experiments either. 
For all our experiments our CNN topologies are based on the ones used in ~\cite{ganin2016domain}, to be comparable to previous work in unsupervised domain adaptation. The exact architectures for all models are shown in our Supplementary Material. 

 In our framework, CORAL \cite{sun2015return} would be equivalent to fixing our shared representation matrices $\bs H_c^s$ and $\bs H_c^t$, normalizing them and then minimizing $ \| \bs A{\bs H_c^s}^\top {\bs H_c^s}\bs A^\top - {\bs H_c^t}^\top {\bs H_c^t} \|_F^2$ with respect to a weight matrix $\bs A$ that aligns the two correlation matrices. For the {CORAL} experiments, we follow the suggestions of~\cite{sun2015return}, and extract features for both source and target domains from the penultimate layer of each network. Once the correlation matrices for each domain are aligned, we evaluate on the target test data the performance of a linear support vector machine (SVM) classifier trained on the source training data. The SVM penalty parameter was optimized based on the target domain validation set for each of our domain adaptation scenarios. 
For MMD regularization, we used a linear combination of 19 RBF kernels\footnote{The Supplementary Material has details on all the parameters.}. We applied MMD on \textsl{fc3} on all our model architectures and minimized
$\mathcal {L}=\mathcal {L}_\mathrm{class}+\gamma\; \mathcal {L}_\mathrm{similarity}^\mathrm{MMD}$ with respect to $\bs \theta_c,\bs \theta_g$.
Preliminary experiments with having MMD applied on more than one layers did not show any performance improvement for our experiments and architectures. 
For DANN regularization, we applied the GRL and the domain classifier as prescribed in \cite{ganin2016domain} for each scenario. We optimized
$\mathcal {L}=\mathcal {L}_\mathrm{class}+\gamma\; \mathcal\mathcal {L}_\mathrm{similarity}^\mathrm{DANN}$ by minimizing it with respect to $\bs \theta_c,\bs \theta_g$ and maximizing it with respect to the domain classifier parameters $\bs \theta_z$.

For our Domain Separation Network experiments, our similarity losses are always applied at the first fully connected layer of each network after a number of convolutional and max pooling layers. For each private space encoder network we use a simple convolutional and max pooling structure followed by a fully-connected layer with a number of nodes equal to the number of nodes at the final layer $\bs h_c$ of the equivalent shared encoder $E_c$. The output of the shared and private encoders gets added before being fed to the shared decoder $D$.
For the latter we use a deconvolutional architecture \cite{zeiler2010deconvolutional} which consists of a fully connected layer with 300 nodes, a resizing layer to $10\times10\times3$, two $3\times3\times16$ convolutional layers, one upsampling layer to $32\times32\times16$, another $3\times3\times16$ convolutional layer, followed by the reconstruction output.
\subsection{Discussion}
\label{sec:discussion}
 The DSN with DANN model outperforms all the other methods we experimented with for all our unsupervised domain adaptation scenarios (see \autoref{tab:results} and \ref{tab:pose_results}). Our unsupervised domain separation networks are able to improve both upon MMD regularization and DANN.  Using DANN as a similarity loss (\autoref{eq:dann}) worked better than using MMD (\autoref{eq:mmd}) as a similarity loss, which is consistent with results obtained for domain adaptation using MMD regularization and DANN alone.
% \paragraph{The effect of the various components}

In order to examine the effect of the soft orthogonality constraints ($\mathcal{L}_\mathrm{difference}$), we took our best model, our DSN model with the DANN loss, and removed these constraints by setting the $\beta$ coefficient to $0$. Without them, the model performed consistently worse in all scenarios.
We also validated our choice of our scale--invariant mean squared error reconstruction loss as opposed to the more popular mean squared error loss by running our best model with $\mathcal{L}_\mathrm{recon}^{{L2}}=\frac{1}{k}||{\bf x}-\hat{{\bf x}}||_2^2$. With this variation we also get worse classification results consistently, as shown in experiments from \autoref{tab:ablation_results}.
\begin{table}[t]
\centering
\caption{Effect of our difference and reconstruction losses on our best model. The first row is replicated from \autoref{tab:results}. In the second row, we remove the soft orthogonality constraint. In the third row, we replace the scale--invariant MSE with regular MSE.}
\vspace{2mm}
\label{tab:ablation_results}
\begin{tabular}{ | l | l | l | l | l | }
\hline
\bf Model   & \bf MNIST to & \bf Synth. Digits to & \bf SVHN to  & \bf Synth. Signs to \\
 & \bf MNIST-M  & \bf SVHN        & \bf MNIST & \bf GTSRB\\ \hline \hline
All terms  & \textbf{83.23} & \textbf{91.22} & \textbf{82.78} & \textbf{93.01} \\ \hline 
No $\mathcal{L}_\mathrm{difference}$ & 80.26 & 89.21 &  80.54 & 91.89\\ \hline
With $\mathcal{L}_\mathrm{recon}^{L2}$ & 80.42 & 88.98 &  79.45 & 92.11\\ \hline
\end{tabular}
\end{table}

The shared and private representations of each domain are combined for the reconstruction of samples. Individually decoding the shared and private representations gives us reconstructions that serve as useful depictions of our domain adaptation process. In \autoref{fig:mnist_m_visualizations} we use the ``MNIST to MNIST-M''  and the ``Synth. Objects to LINEMOD'' scenarios   for such visualizations.  In the former scenario, the model clearly separates the foreground from the background and produces a shared space that is very similar to the source domain. This is expected since the target is a transformation of the source. In the latter scenario, the model is able to produce visualizations of the shared representation that look very similar between source and target domains, which are useful for classification and pose estimation, as shown in \autoref{tab:pose_results}.

\section{Conclusion}
\label{sec:conclusion}
\vspace{-4mm}
We present in this work a deep learning model that improves upon existing unsupervised domain adaptation techniques. The model does so by explicitly separating representations private to each domain and shared between source and target domains. By using existing domain adaptation techniques to make the shared representations similar, and soft subspace orthogonality constraints to make private and shared representations dissimilar, our method outperforms all existing unsupervised domain adaptation methods in a number of adaptation scenarios that focus on the synthetic--to--real paradigm. %This work has clear extensions to tasks such as image segmentation, where arguably the shared representation might become more powerful. %Another research avenue could be to encode source samples with the target private encoder, add the encodings to the shared representation and use the relevant reconstructions as labeled data.

\subsubsection*{Acknowledgments}

We would like to thank Samy Bengio, Kevin Murphy, and Vincent Vanhoucke for valuable comments on this work. We would also like to thank Yaroslav Ganin and Paul Wohlhart for providing some of the datasets we used.

\clearpage
{
\small
\bibliographystyle{ieee}
\bibliography{domain_separation}

\begin{thebibliography}{10}\itemsep=-1pt

\bibitem{abadi2016tensorflow}
M.~Abadi et~al.
\newblock Tensorflow: Large-scale machine learning on heterogeneous distributed
  systems.
\newblock {\em Preprint arXiv:1603.04467}, 2016.

\bibitem{Ajakan2014}
H.~Ajakan, P.~Germain, H.~Larochelle, F.~Laviolette, and M.~Marchand.
\newblock Domain-adversarial neural networks.
\newblock In {\em Preprint, http://arxiv.org/abs/1412.4446}, 2014.

\bibitem{arbelaez2011contour}
P.~Arbelaez, M.~Maire, C.~Fowlkes, and J.~Malik.
\newblock Contour detection and hierarchical image segmentation.
\newblock {\em TPAMI}, 33(5):898--916, 2011.

\bibitem{ben2010theory}
S.~Ben-David, J.~Blitzer, K.~Crammer, A.~Kulesza, F.~Pereira, and J.~W.
  Vaughan.
\newblock A theory of learning from different domains.
\newblock {\em Machine learning}, 79(1-2):151--175, 2010.

\bibitem{caseiro2015beyond}
R.~Caseiro, J.~F. Henriques, P.~Martins, and J.~Batist.
\newblock {Beyond the shortest path: Unsupervised Domain Adaptation by Sampling
  Subspaces Along the Spline Flow}.
\newblock In {\em CVPR}, 2015.

\bibitem{eigen2014depth}
D.~Eigen, C.~Puhrsch, and R.~Fergus.
\newblock Depth map prediction from a single image using a multi-scale deep
  network.
\newblock In {\em NIPS}, pages 2366--2374, 2014.

\bibitem{ganin2014unsupervised}
Y.~Ganin and V.~Lempitsky.
\newblock Unsupervised domain adaptation by backpropagation.
\newblock In {\em ICML}, pages 513--520, 2015.

\bibitem{ganin2016domain}
Y.~Ganin~\etal.
\newblock {Domain-Adversarial Training of Neural Networks}.
\newblock {\em JMLR}, 17(59):1--35, 2016.

\bibitem{gong2012geodesic}
B.~Gong, Y.~Shi, F.~Sha, and K.~Grauman.
\newblock Geodesic flow kernel for unsupervised domain adaptation.
\newblock In {\em CVPR}, pages 2066--2073. IEEE, 2012.

\bibitem{gopalan2011domain}
R.~Gopalan, R.~Li, and R.~Chellappa.
\newblock {Domain Adaptation for Object Recognition: An Unsupervised Approach}.
\newblock In {\em ICCV}, 2011.

\bibitem{gretton2012mmd}
A.~Gretton, K.~M. Borgwardt, M.~J. Rasch, B.~Sch{\"o}lkopf, and A.~Smola.
\newblock {A Kernel Two-Sample Test}.
\newblock {\em JMLR}, pages 723--773, 2012.

\bibitem{griffin2007caltech}
G.~Griffin, A.~Holub, and P.~Perona.
\newblock Caltech-256 object category dataset.
\newblock {\em CNS-TR-2007-001}, 2007.

\bibitem{hinterstoisser2012accv}
S.~Hinterstoisser~\etal.
\newblock Model based training, detection and pose estimation of texture-less
  3d objects in heavily cluttered scenes.
\newblock In {\em ACCV}, 2012.

\bibitem{huynh2009metrics}
D.~Q. Huynh.
\newblock Metrics for 3d rotations: Comparison and analysis.
\newblock {\em Journal of Mathematical Imaging and Vision}, 35(2):155--164,
  2009.

\bibitem{jia2010factorized}
Y.~Jia, M.~Salzmann, and T.~Darrell.
\newblock Factorized latent spaces with structured sparsity.
\newblock In {\em NIPS}, pages 982--990, 2010.

\bibitem{lecun1998gradient}
Y.~LeCun, L.~Bottou, Y.~Bengio, and P.~Haffner.
\newblock Gradient-based learning applied to document recognition.
\newblock {\em Proceedings of the IEEE}, 86(11):2278--2324, 1998.

\bibitem{lin2014microsoft}
T.-Y. Lin, M.~Maire, S.~Belongie, J.~Hays, P.~Perona, D.~Ramanan,
  P.~Doll{\'a}r, and C.~L. Zitnick.
\newblock Microsoft coco: Common objects in context.
\newblock In {\em ECCV 2014}, pages 740--755. Springer, 2014.

\bibitem{long2015learning}
M.~Long and J.~Wang.
\newblock Learning transferable features with deep adaptation networks.
\newblock {\em ICML}, 2015.

\bibitem{mansour2009domain}
Y.~Mansour~\etal.
\newblock {Domain adaptation with multiple sources}.
\newblock In {\em NIPS}, 2009.

\bibitem{Moiseev2013}
B.~Moiseev, A.~Konev, A.~Chigorin, and A.~Konushin.
\newblock {\em Evaluation of Traffic Sign Recognition Methods Trained on
  Synthetically Generated Data}, chapter ACIVS, pages 576--583.
\newblock Springer International Publishing, 2013.

\bibitem{netzer2011reading}
Y.~Netzer, T.~Wang, A.~Coates, A.~Bissacco, B.~Wu, and A.~Y. Ng.
\newblock Reading digits in natural images with unsupervised feature learning.
\newblock In {\em {NIPS Workshops}}, 2011.

\bibitem{olson2011apriltag}
E.~Olson.
\newblock Apriltag: A robust and flexible visual fiducial system.
\newblock In {\em Robotics and Automation (ICRA), 2011 IEEE International
  Conference on}, pages 3400--3407. IEEE, 2011.

\bibitem{ILSVRC15}
O.~Russakovsky et~al.
\newblock {ImageNet Large Scale Visual Recognition Challenge}.
\newblock {\em IJCV}, 115(3):211--252, 2015.

\bibitem{saenko2010adapting}
K.~Saenko~\etal.
\newblock Adapting visual category models to new domains.
\newblock In {\em ECCV}. Springer, 2010.

\bibitem{salzmann2010factorized}
M.~Salzmann~et. al.
\newblock Factorized orthogonal latent spaces.
\newblock In {\em AISTATS}, pages 701--708, 2010.

\bibitem{stallkamp2012gtsrb}
J.~Stallkamp, M.~Schlipsing, J.~Salmen, and C.~Igel.
\newblock Man vs. computer: Benchmarking machine learning algorithms for
  traffic sign recognition.
\newblock {\em Neural Networks}, 2012.

\bibitem{sun2015return}
B.~Sun, J.~Feng, and K.~Saenko.
\newblock Return of frustratingly easy domain adaptation.
\newblock In {\em AAAI}. 2016.

\bibitem{sutskever2013momentum}
I.~Sutskever, J.~Martens, G.~Dahl, and G.~Hinton.
\newblock On the importance of initialization and momentum in deep learning.
\newblock In {\em ICML}, pages 1139--1147, 2013.

\bibitem{tzeng2015simultaneous}
E.~Tzeng, J.~Hoffman, T.~Darrell, and K.~Saenko.
\newblock {Simultaneous deep transfer across domains and tasks}.
\newblock In {\em CVPR}, pages 4068--4076, 2015.

\bibitem{tzeng2015ddc}
E.~Tzeng, J.~Hoffman, N.~Zhang, K.~Saenko, and T.~Darrell.
\newblock Deep domain confusion: Maximizing for domain invariance.
\newblock {\em Preprint arXiv:1412.3474}, 2014.

\bibitem{virtanen2011bayesian}
S.~Virtanen, A.~Klami, and S.~Kaski.
\newblock {Bayesian CCA via group sparsity}.
\newblock In {\em ICML}, pages 457--464, 2011.

\bibitem{wohlhart2015learning}
P.~Wohlhart and V.~Lepetit.
\newblock Learning descriptors for object recognition and 3d pose estimation.
\newblock In {\em CVPR}, pages 3109--3118, 2015.

\bibitem{zeiler2010deconvolutional}
M.~D. Zeiler, D.~Krishnan, G.~W. Taylor, and R.~Fergus.
\newblock Deconvolutional networks.
\newblock In {\em CVPR}, pages 2528--2535. IEEE, 2010.

\end{thebibliography}
}

% Uncomment this to include the supplementary in the
% main paper.
\clearpage
\section*{Supplementary Material}
\appendix 
% \graphicspath{{gfx/supp/}}

%As mentioned in \autoref{sec:related_work} existing work around unsupervised domain adaptation for neural networks learns to extract the shared information between the two domains, by using either by matching moments of the representations [CORAL; MMD] or by using a discriminant classifier [GRL], which is then used as features for the task-at-hand.
%
%Key points:
%
%\begin{enumerate}
%    \item In the presence of highly correlated noise between the two domains.
%    \item It is crucial to be able to debug the model; visualize the the shared representation.
%\end{enumerate}
%
%To this end, we use ideas from private-shared space models~\cite{jia2010factorized,salzmann2010factorized,virtanen2011bayesian} to jointly model both the private and shared representations of the domains.
%
%By imposing a natural incoherence condition on the representations, which can be viewed as a soft orthogonality constraint~\cite{salzmann2010factorized}, limits the redundancy amongst the shared and private components. We show that these constrains make the problem of finding the underlying distribution amongst the domains makes learning well-posed. Incoherence constraints were used before for a lot of component analysis problems such as Independent Component Analysis (ICA)~
% \cite{le2011ica}%(Quac Le) ...
\section{Correlation Regularization}
\label{sec:correg}

Correlation Alignment (CORAL)~\cite{sun2015return} aims to find a mapping from the representations of the source domain to the representations of the target domain by matching only the second--order statistics. In our framework, this would be equivalent to fixing our common representation  matrices $\bs H_c^s$ and $\bs H_c^t$ after normalizing them and then finding a weight matrix $ \hat{\bs A} = \displaystyle {\argmin_{\bs A}}\left\| \bs A{\bs H_c^s}^\top {\bs H_c^s}\bs A^\top - {\bs H_c^t}^\top {\bs H_c^t} \right\|_F^2$ that aligns the two correlation matrices. Although this has the advantage that the optimization
is convex and can be solved in closed form, all convolutional features remain
fixed during the process, which might not be optimal for the task at hand. Also, because of this we are not able to use it as a similarity loss for our DSNs. 
Motivated by this shortcoming, we propose here a new domain adaptation method, Correlation Regularization (CorReg). We show in \autoref{tab:results} that our new domain adaptation method, which is theoretically as powerful as an MMD loss with a second--order polynomial kernel, outperforms CORAL in all our datasets. Adapting a feature hierarchy to be domain--invariant is more powerful than learning a mapping from the representations of one domain to those of another. Moreover, we use it as yet another similarity loss for our Domain Separation Networks:
\begin{equation}
\mathcal{L}_\mathrm{similarity}^\mathrm{CorReg} = \left\| {\bs H_c^s}^\top {\bs H_c^s} - {\bs H_c^t}^\top {\bs H_c^t} \right\|_F^2
\end{equation}
Our DNS with CorReg performs better than both CORAL and CorReg, which is consistent with the rest of our results.

\begin{table}[h]
\centering
\caption{Our main results from the paper with two additional lines for CorReg and DSN with CorReg.}
\label{tab:results}
\begin{tabular}{ | l | l | l | l | l | }
\hline
\bf Model   & \bf MNIST to & \bf Synth Digits to &\bf SVHN to  &\bf Synth Signs to \\
 &\bf MNIST-M  &\bf SVHN        &\bf MNIST &\bf GTSRB     \\ \hline \hline
Source-only  & 56.6 (52.2) & 86.7 (86.7)      & 59.2 (54.9) & 85.1  (79.0)    \\ \hline \hline
CORAL \cite{sun2015return} & 57.7 & 85.2       &  63.1     & 86.9        \\ \hline
CorReg (Ours)  & 62.06 & 87.33       & 69.20 & 90.75     \\ \hline\hline
MMD  \cite{tzeng2015ddc,long2015learning}  & 76.9 & 88.0 & 71.1 & 91.1 \\ \hline 
DANN \cite{ganin2016domain}  & 77.4 (76.6)  &  90.3 (91.0)    & 70.7 (73.8)  &     92.9 (88.6) \\ \hline
DSN w/ MMD (ours)  & 80.5 & 88.5  & 72.2   & 92.6 \\ \hline
DSN w/ DANN (ours) & \textbf{83.2} & \textbf{91.2} & \textbf{82.7} & \textbf{93.1} \\ \hline\hline
Target-only  & 98.7 & 92.4  & 99.5 & 99.8  \\ \hline
\end{tabular}
\end{table}

\section{Office Dataset Criticism}
\begin{figure}[th]
    \centering
    \begin{subfigure}[b]{0.18\linewidth}
        \centering
        \includegraphics[width=\linewidth]{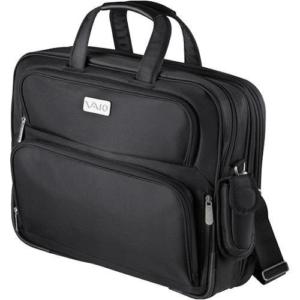}
    \end{subfigure}    \hfill
        \begin{subfigure}[b]{0.18\linewidth}
        \centering
        \includegraphics[width=\linewidth]{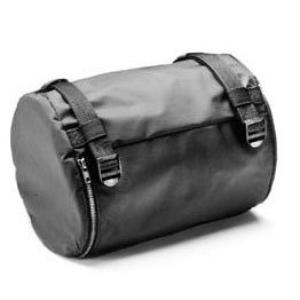}
    \end{subfigure}    \hfill
        \begin{subfigure}[b]{0.18\linewidth}
        \centering
        \includegraphics[width=\linewidth]{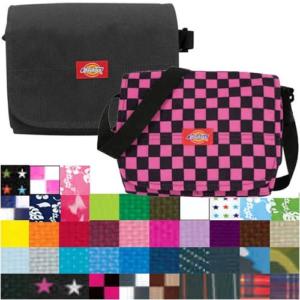}
    \end{subfigure}    \hfill
        \begin{subfigure}[b]{0.18\linewidth}
        \centering
        \includegraphics[width=\linewidth]{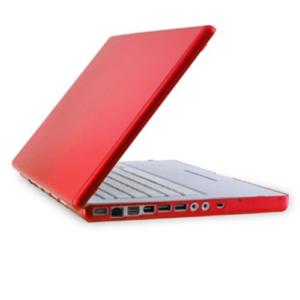}
    \end{subfigure}   \hfill
    \begin{subfigure}[b]{0.18\linewidth}
        \centering
        \includegraphics[width=\linewidth]{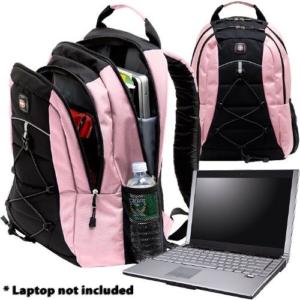}
    \end{subfigure}  \\    
    \begin{subfigure}[b]{0.18\linewidth}
        \centering
        \includegraphics[width=\linewidth]{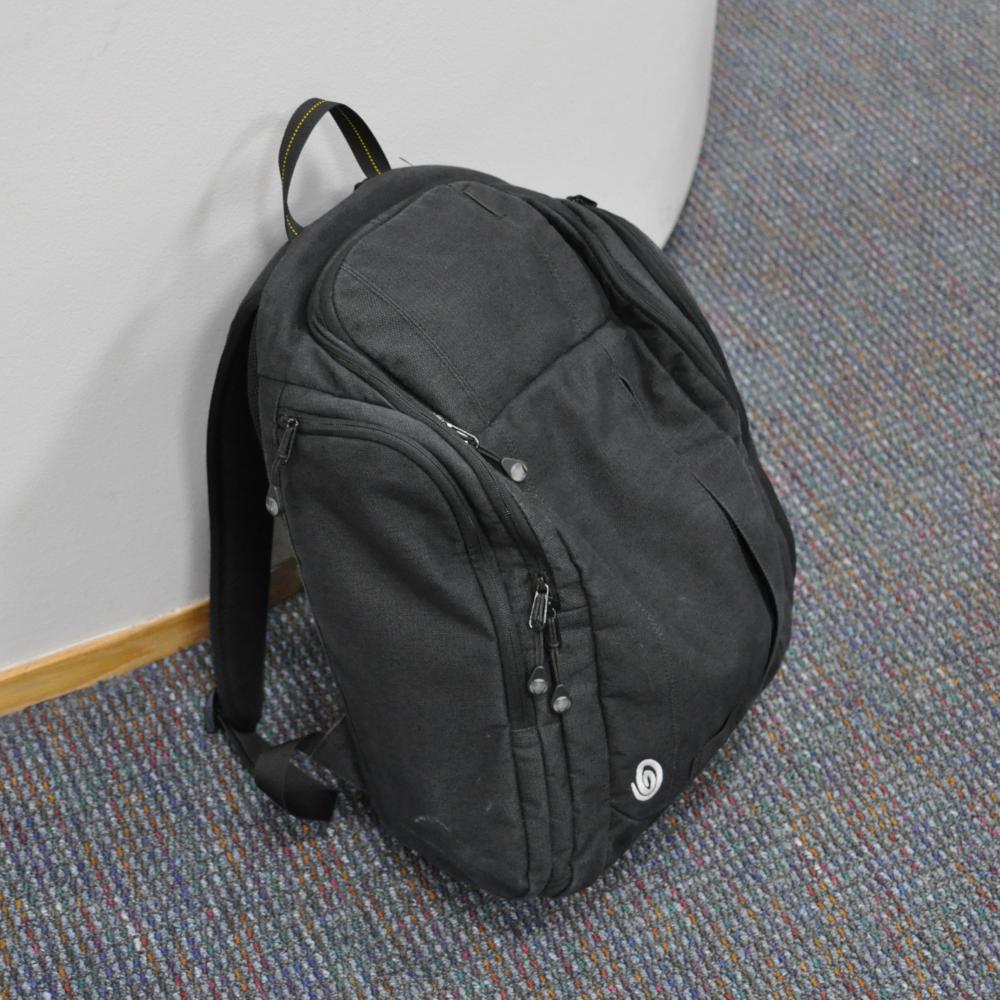}
    \end{subfigure}    \hfill
        \begin{subfigure}[b]{0.18\linewidth}
        \centering
        \includegraphics[width=\linewidth]{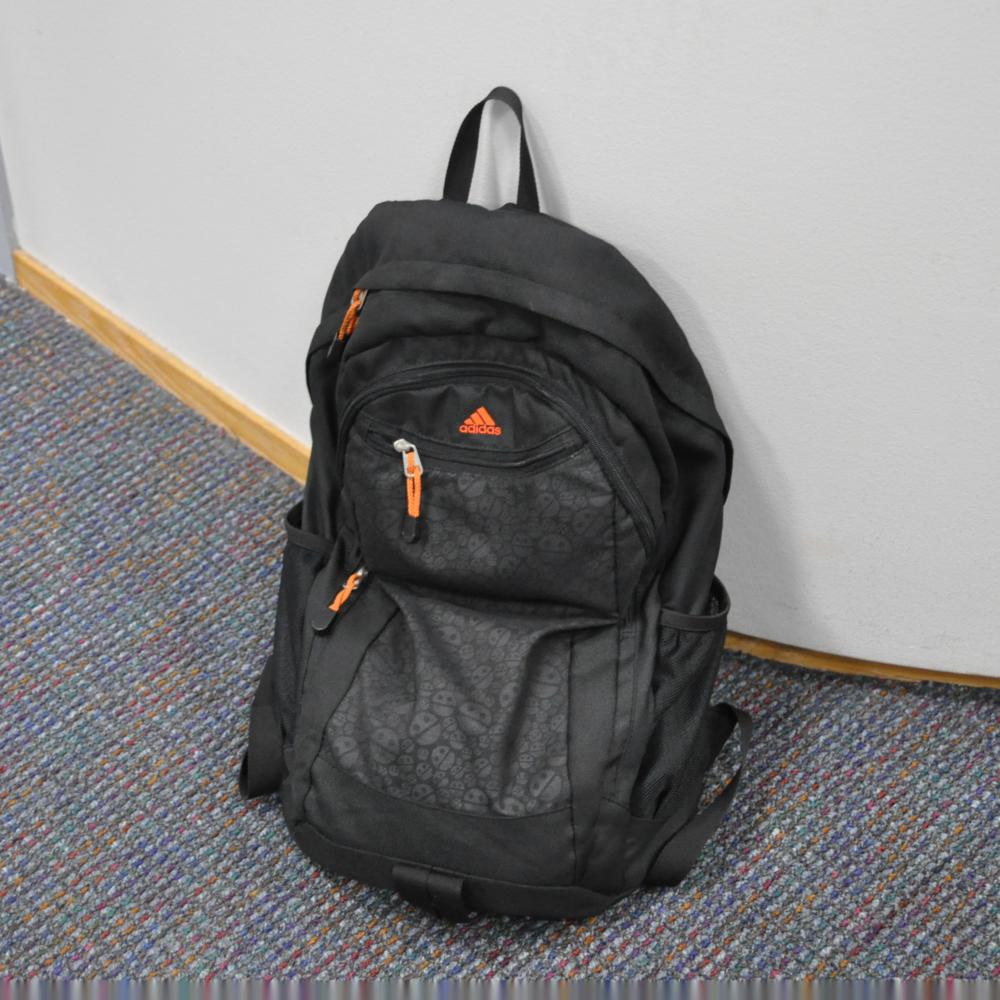}
    \end{subfigure}    \hfill
        \begin{subfigure}[b]{0.18\linewidth}
        \centering
        \includegraphics[width=\linewidth]{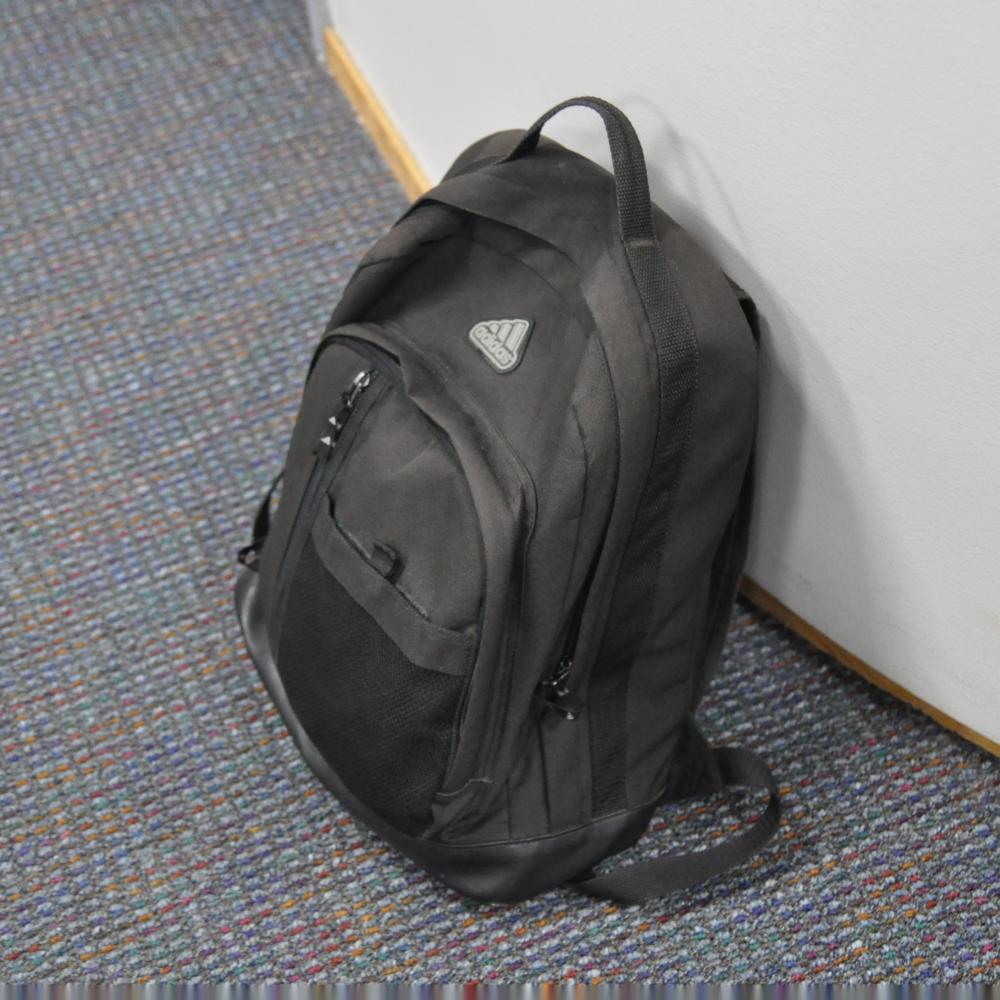}
    \end{subfigure}    \hfill
        \begin{subfigure}[b]{0.18\linewidth}
        \centering
        \includegraphics[width=\linewidth]{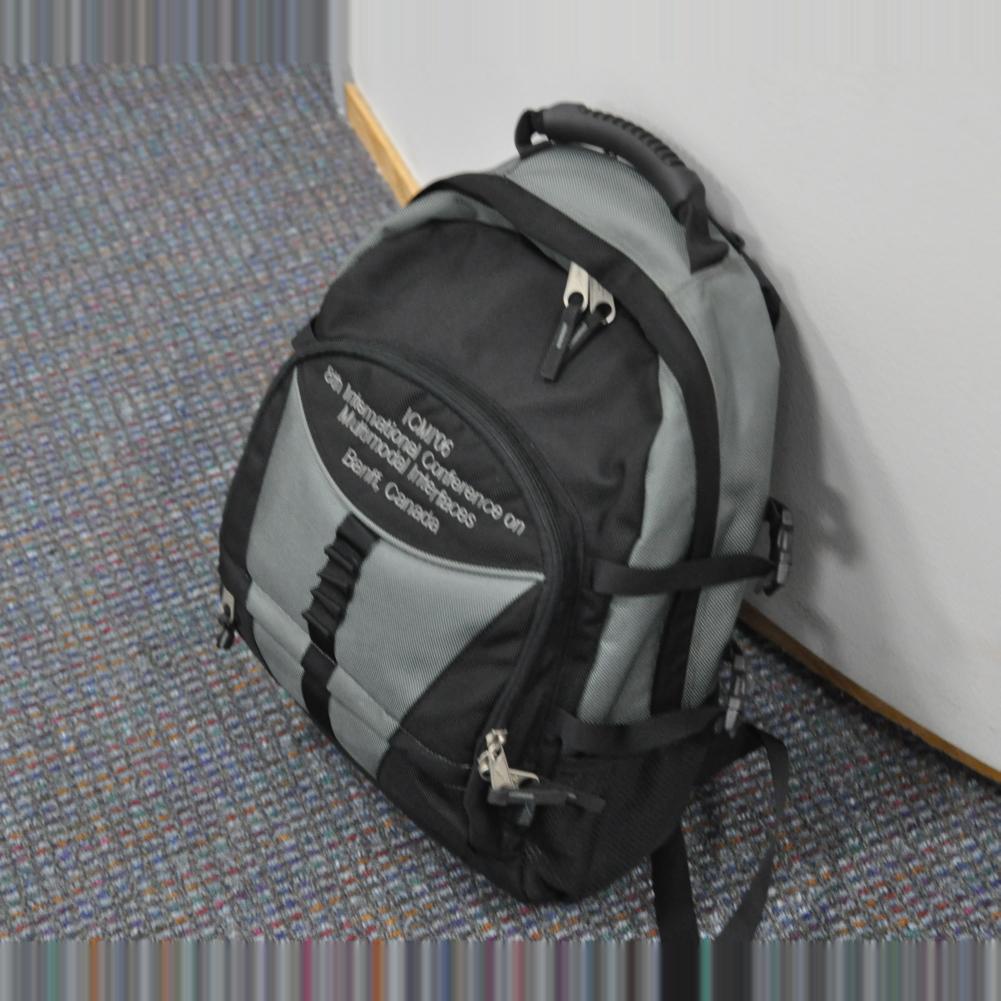}
    \end{subfigure}   \hfill
    \begin{subfigure}[b]{0.18\linewidth}
        \centering
        \includegraphics[width=\linewidth]{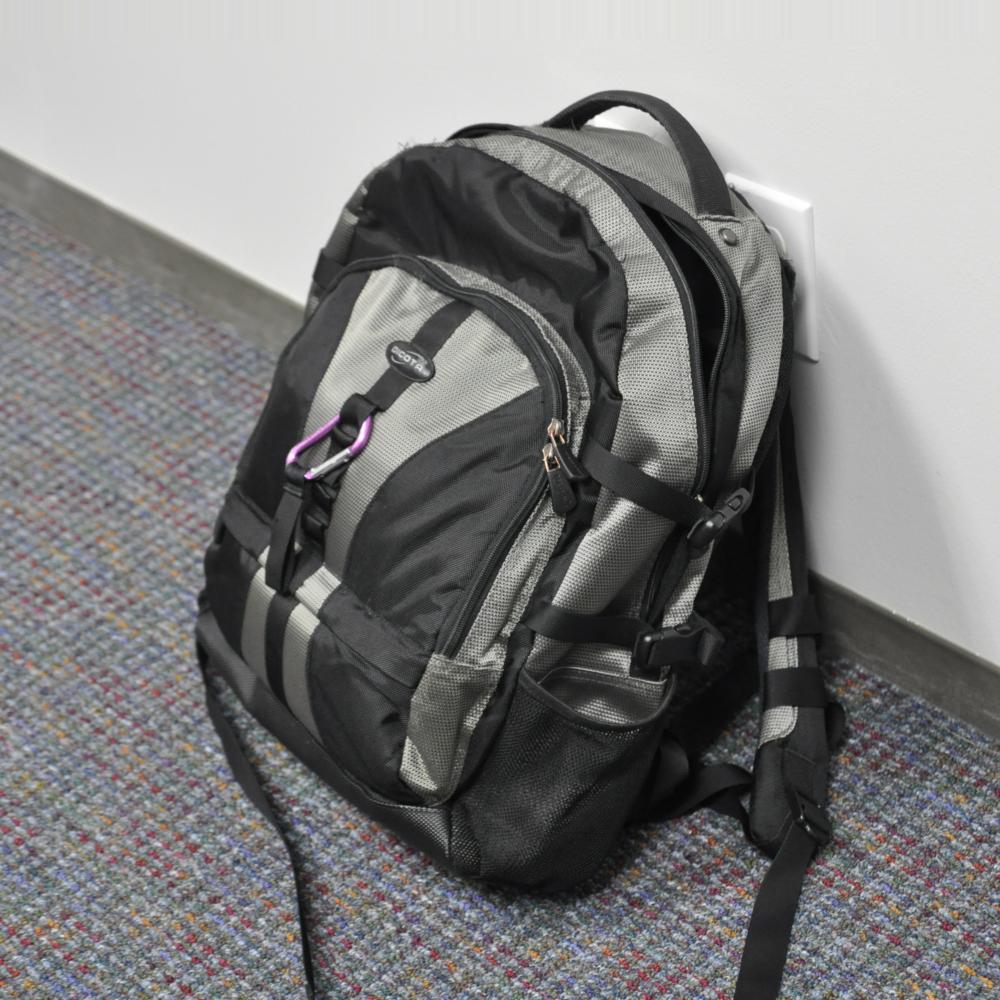}
    \end{subfigure}  \\
    \begin{subfigure}[b]{0.18\linewidth}
        \centering
        \includegraphics[width=\linewidth]{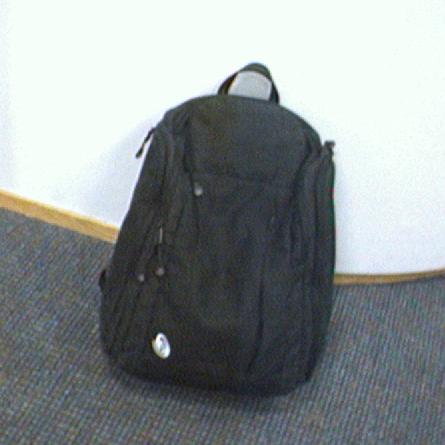}
    \end{subfigure}    \hfill
        \begin{subfigure}[b]{0.18\linewidth}
        \centering
        \includegraphics[width=\linewidth]{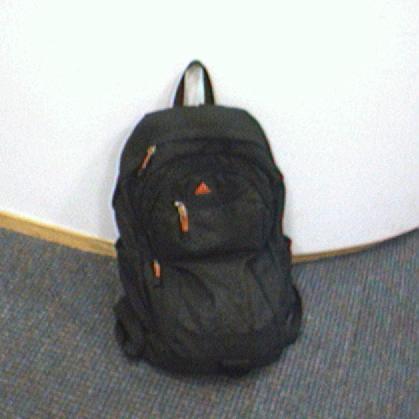}
    \end{subfigure}    \hfill
        \begin{subfigure}[b]{0.18\linewidth}
        \centering
        \includegraphics[width=\linewidth]{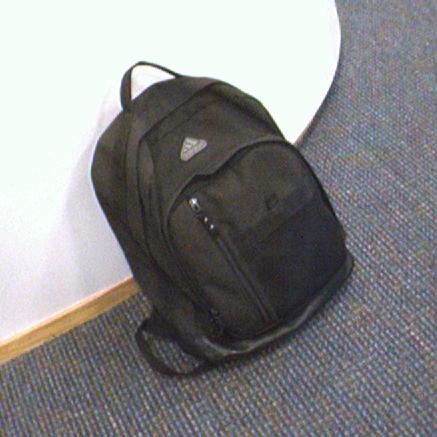}
    \end{subfigure}    \hfill
        \begin{subfigure}[b]{0.18\linewidth}
        \centering
        \includegraphics[width=\linewidth]{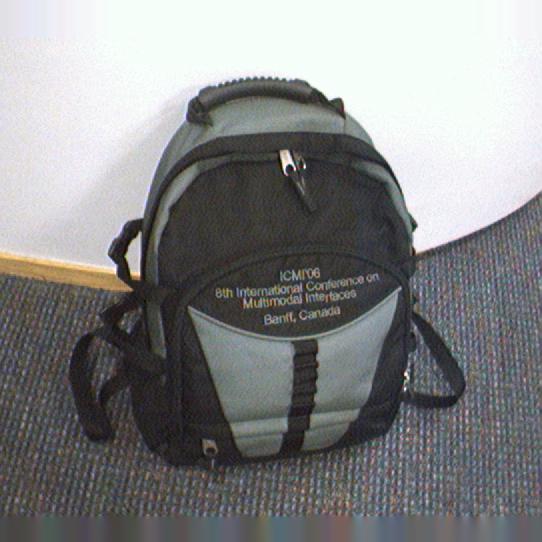}
    \end{subfigure}   
    \begin{subfigure}[b]{0.18\linewidth}
        \centering
        \includegraphics[width=\linewidth]{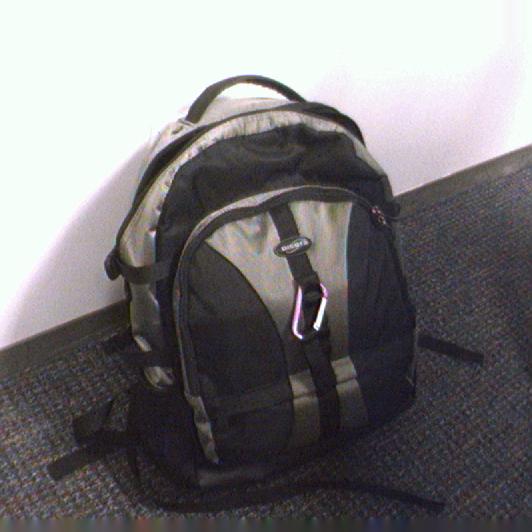}
    \end{subfigure}  \\
    \begin{subfigure}[b]{0.18\linewidth}
        \centering
        \includegraphics[width=\linewidth]{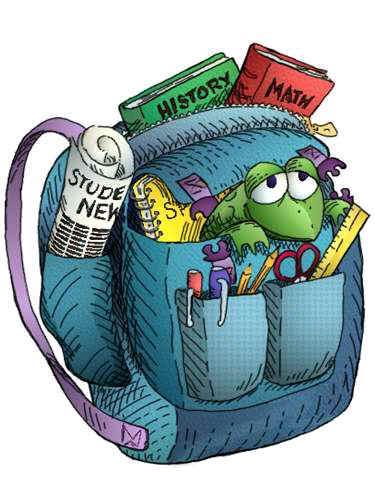}
    \end{subfigure}    \hfill
        \begin{subfigure}[b]{0.18\linewidth}
        \centering
        \includegraphics[width=\linewidth]{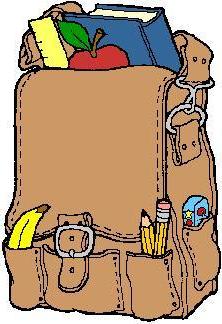}
    \end{subfigure}    \hfill
        \begin{subfigure}[b]{0.18\linewidth}
        \centering
        \includegraphics[width=\linewidth]{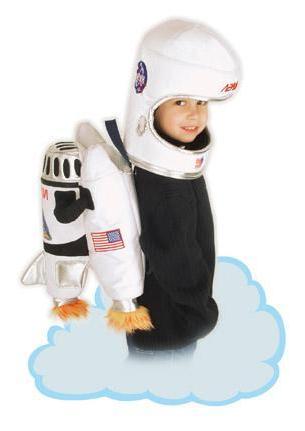}
    \end{subfigure}    \hfill
        \begin{subfigure}[b]{0.18\linewidth}
        \centering
        \includegraphics[width=\linewidth]{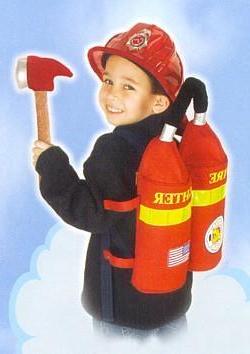}
    \end{subfigure}   
    \begin{subfigure}[b]{0.18\linewidth}
        \centering
        \includegraphics[width=\linewidth]{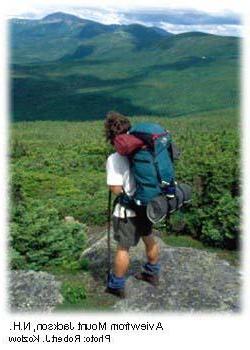}
    \end{subfigure}      
    \caption{Examples of the `back\_pack' class in the different domains in Office and Caltech--256. \textbf{First Row:} 5 of the 92 images in the Amazon domain. \textbf{Second Row:} The DSLR domain contains 4 images for the rightmost image from different frontal angles, 2 images for the other 4 backpacks for a total of 12 images for this class. \textbf{Third Row:} The webcam domain contains the exact same backpacks with DSLR with similar poses for a total of 29 images for this class. \textbf{Fourth Row:} Some of the 151 backpack samples Caltech domain.}
    \label{fig:office_pollution2}
\end{figure}

The most commonly used dataset for visual domain adaptation in the context of object classification is Office \cite{saenko2010adapting}, sometimes combined with the Caltech--256 dataset \cite{griffin2007caltech} as an additional domain. However, these datasets exhibit significant variations in both low-level and high-level parameter distributions. Low-level variations are due to the different cameras and background textures in the images (e.g. Amazon versus DSLR), which is welcome. However, there are significant high-level variations due to elements like label pollution: e.g. the motorcycle class contains non-motorcycle objects; the backpack class contains 2 laptops; some classes contain the object in only one pose. Other commonly used datasets such as Caltech-256 suffer from similar problems. We illustrate some of these issues for the `back\_pack' class for its 92 Amazon samples, its 12 DSLR samples, its 29 Webcam samples, and its 151 Caltech samples in \autoref{fig:office_pollution2}. Other classes exhibit similar problems. For these reasons some works, eg \cite{sun2015return}, pretrain their models on Imagenet before performing the domain adaptation in these scenarios. This essentially involves another source domain (Imagenet) in the transfer.

\section{Domain Separation}
We visualize in~\autoref{fig:sdads} reconstructions for both source and target domains of each domain adaptation scenario. Although the visualizations are not as clear as with the ``MNIST to MNIST-M'' scenario, where the target domain was a direct transformation of the source domain, it is interesting to note the similarities of the visualizations of the shared representations, and the exclusion of some shared information in the private domains.

\newcommand{\subfigscale}{0.24}

\begin{figure}[ht]
    \centering
    
    \begin{subfigure}[b]{\subfigscale\linewidth}
        \centering
        \includegraphics[width=\linewidth]{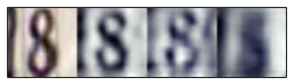}
    \end{subfigure}
    \begin{subfigure}[b]{\subfigscale\linewidth}
        \centering
        \includegraphics[width=\linewidth]{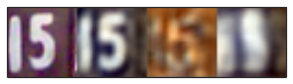}
    \end{subfigure}
    \begin{subfigure}[b]{\subfigscale\linewidth}
        \centering
        \includegraphics[width=\linewidth]{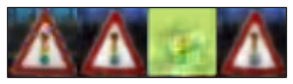}
    \end{subfigure}
    \begin{subfigure}[b]{\subfigscale\linewidth}
        \centering
        \includegraphics[width=\linewidth]{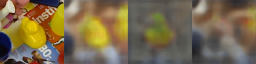}
    \end{subfigure} \\
    \begin{subfigure}[b]{\subfigscale\linewidth}
        \centering
        \includegraphics[width=\linewidth]{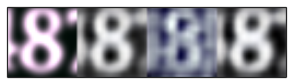}
        \caption{}
    \end{subfigure}
    \begin{subfigure}[b]{\subfigscale\linewidth}
        \centering
        \includegraphics[width=\linewidth]{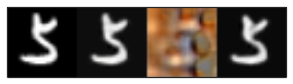}
        \caption{}
    \end{subfigure}
    \begin{subfigure}[b]{\subfigscale\linewidth}
        \centering
        \includegraphics[width=\linewidth]{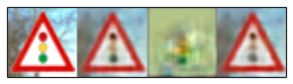}
        \caption{}
    \end{subfigure}
    \begin{subfigure}[b]{\subfigscale\linewidth}
        \centering
        \includegraphics[width=\linewidth]{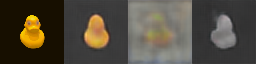}
        \caption{}
    \end{subfigure}
    \caption{
    Reconstructions for the representations of the two domains for \textit{a) Synthetic Digits to SVHN, \textit{b)} SVHN to MNIST, \textit{c)} Synthetic Signs to GTSRB, \textit{d)} Synthetic Objects to LineMOD. In each block from left to right: the original image $\bs x_t$; reconstructed image $D({E_c(\bs x^t) + E_p(\bs x^t)}) $; shared only reconstruction  $D(E_c(\bs x^t)) $;  private only reconstruction  $D(E_p(\bs x^t))$.Reconstructions of target \textit{(top row)} and source \textit{(bottom row)} domains. }.
    }
    \label{fig:sdads}
\end{figure}

\section{Network Topologies and Optimal Parameters}
\begin{figure}
    \centering
    \includegraphics[width=.7\linewidth]{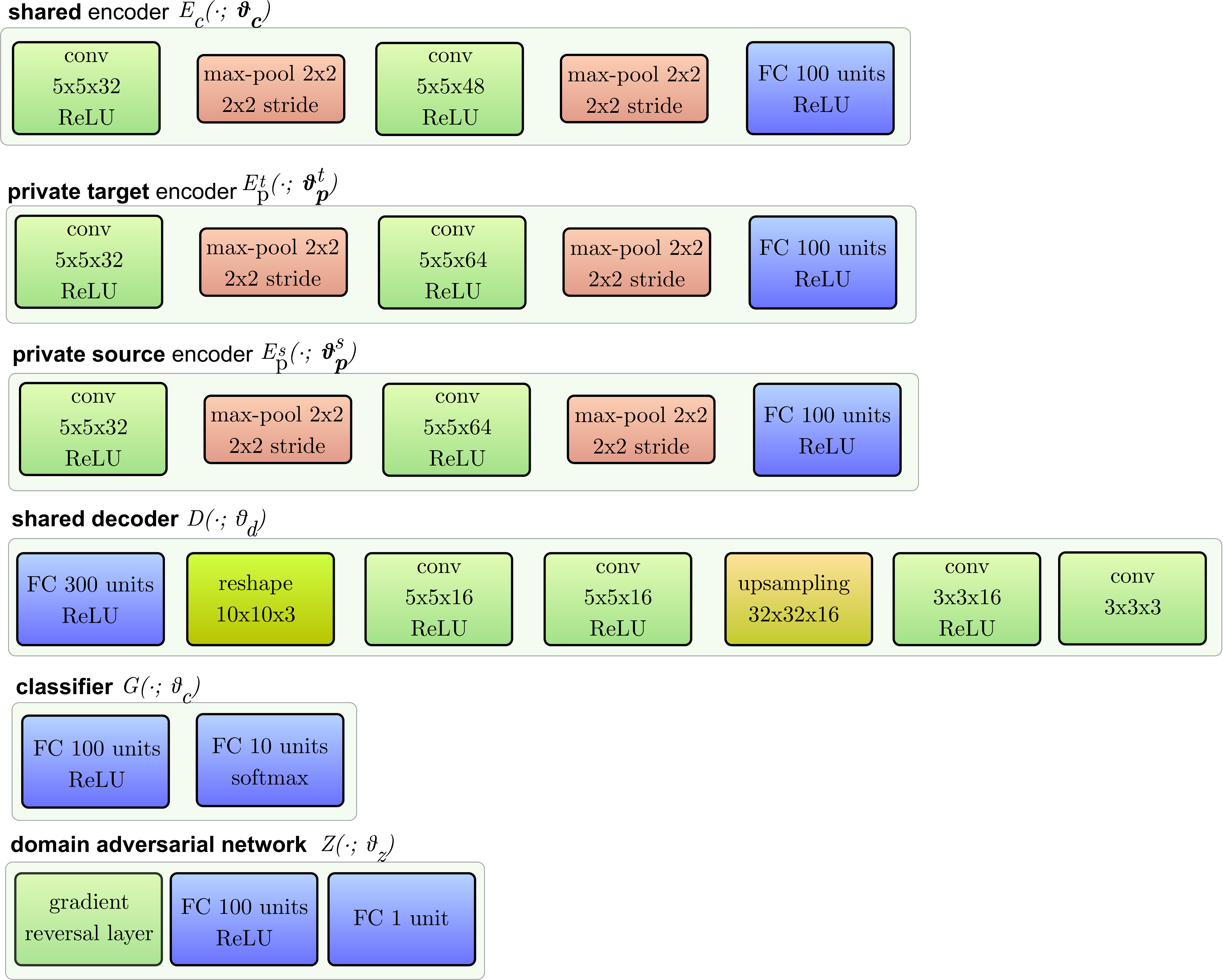}
    \caption{The network topology for ``MNIST to MNIST-M''}
    \label{fig:mnist_arch}
\end{figure}

\begin{figure}
    \centering
    \includegraphics[width=.7\linewidth]{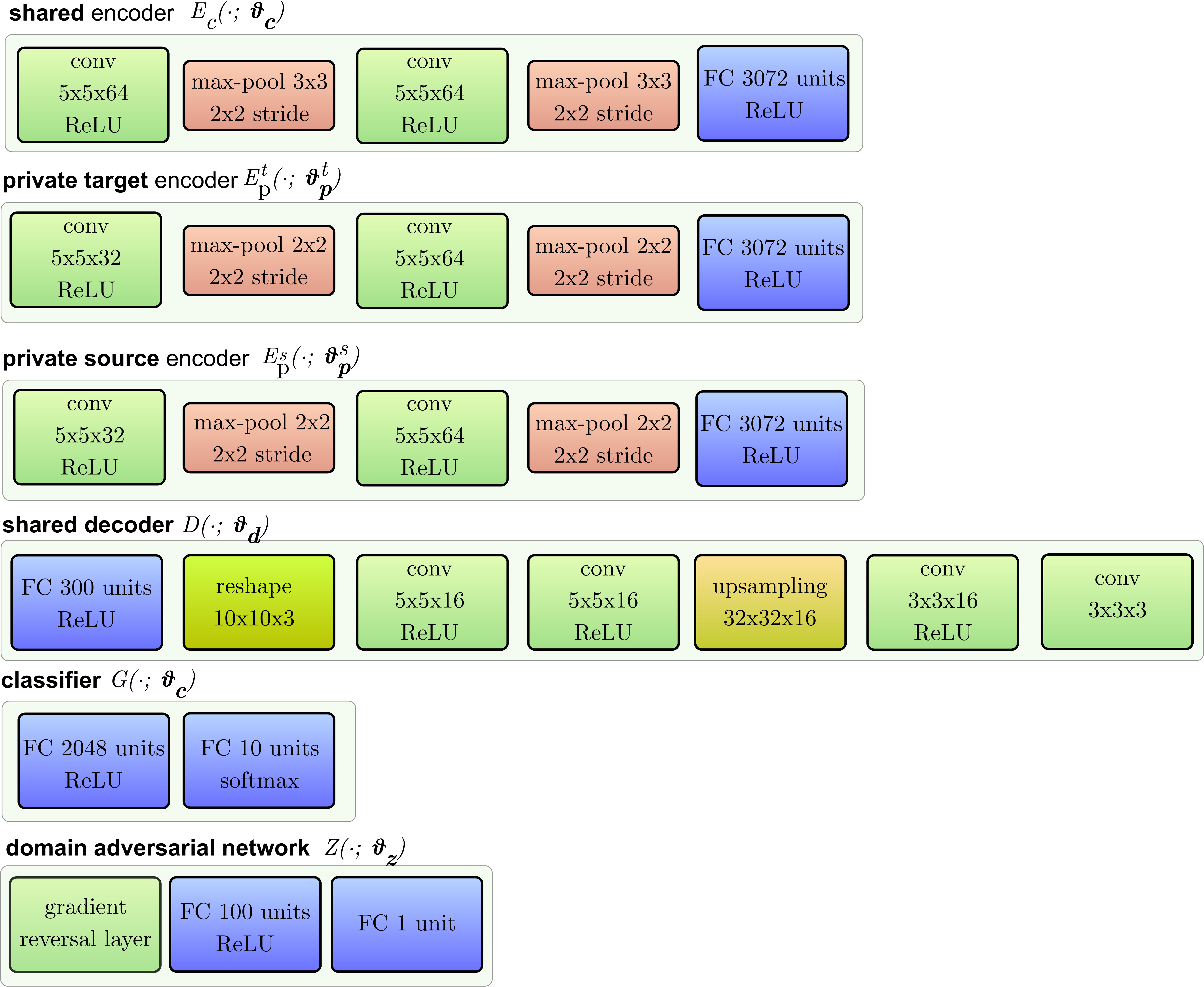}
    \caption{The network topology for ``Synth SVHN to SVHN'' and ``SVHN to MNIST'' experiments.}
    \label{fig:mnist_arch2}
\end{figure}

\begin{figure}
    \centering
    \includegraphics[width=.7\linewidth]{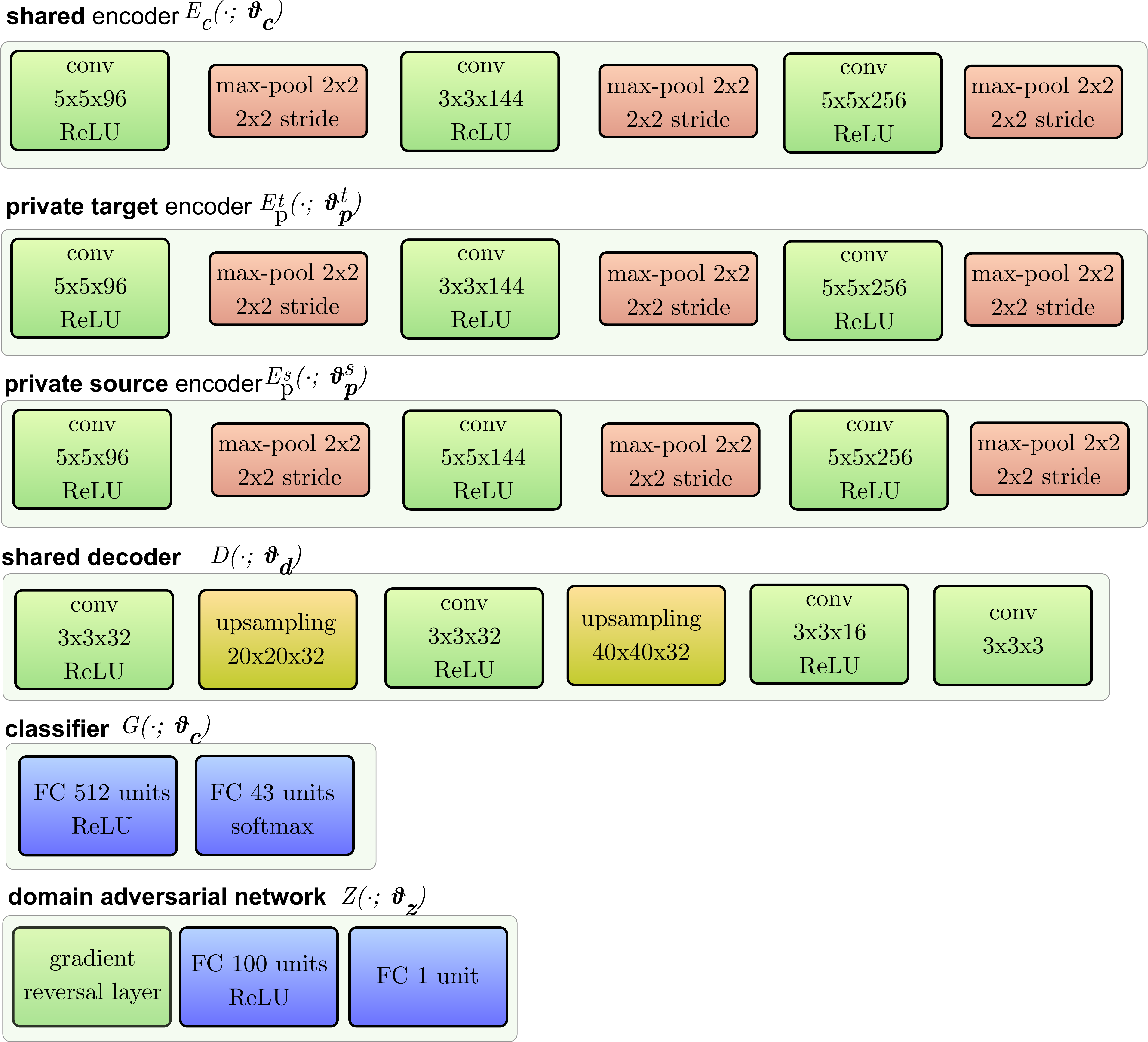}
    \caption{The network topology for ``Synth Signs to GTSRB''}
    \label{fig:mnist_arch3}
\end{figure}

\begin{figure}
    \centering
    \includegraphics[width=.7\linewidth]{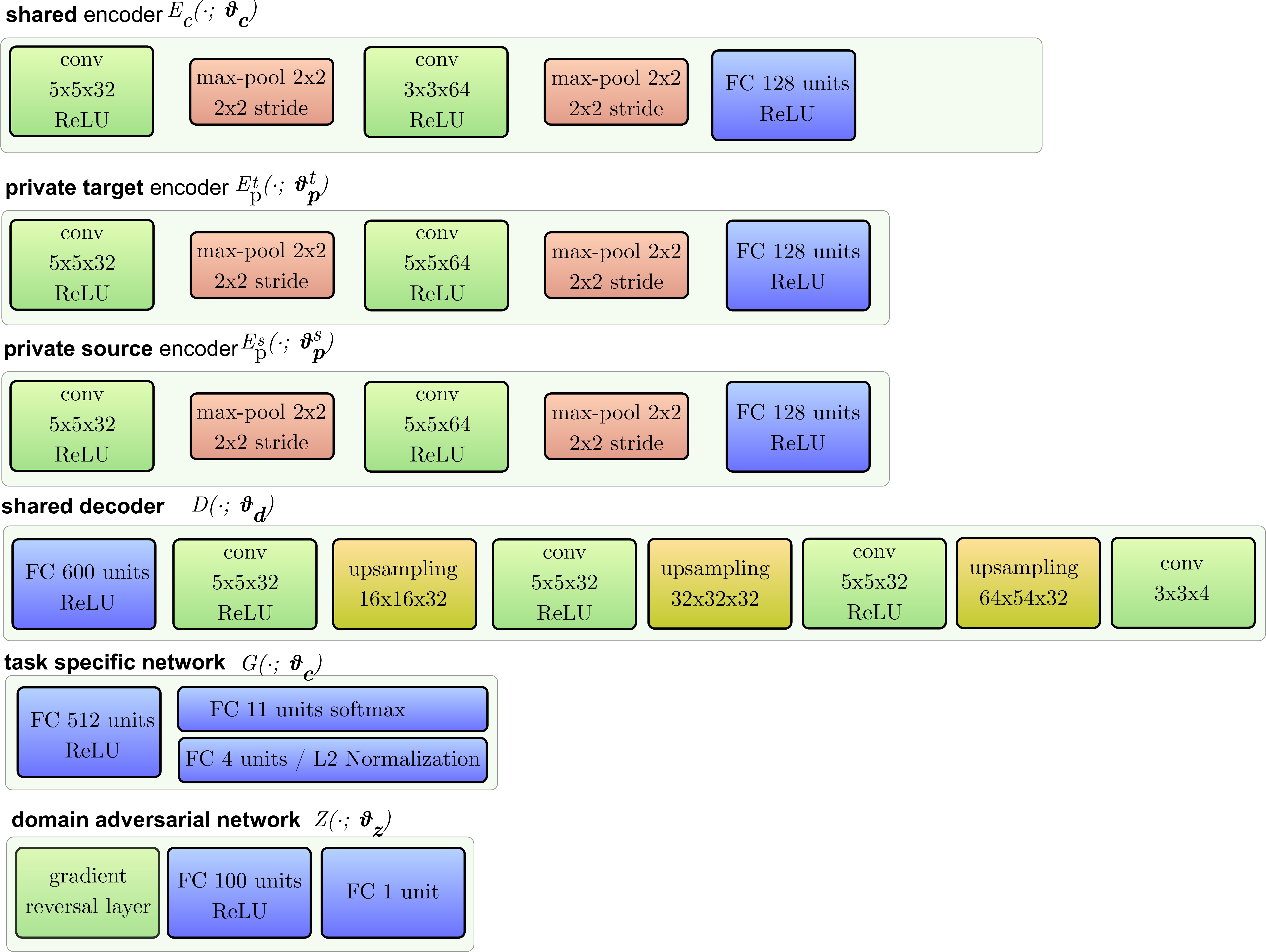}
    \caption{The network topology for ``Synthetic Objects to Linemod''}
    \label{fig:mnist_arch4}
\end{figure}

Since we used different network topologies for our domain adaptation scenarios, there was not enough space to include these in the main paper. We present the exact topologies used in Figures \ref{fig:mnist_arch}--\ref{fig:mnist_arch4}. 

Similarly, we list here all hyperparameters that are important for total reproducibility of all our results.
For CORAL, the SVM
penalty parameter that was optimized based on the validation set for each of our domain adaptation scenarios: $1e^{-4}$ for ``MNIST to MNIST-M'', ``Synth Digits to SVHN'', ``Synth Signs to GTSRB'', and  $1e^{-3}$ for ``SVHN to MNIST''.
For MMD we use 19 RBF kernels with the following standard deviation parameters:
\begin{equation*}
    \bs \sigma = [10^{-6}, 10^{-5}, 10^{-4}, 10^{-3}, 10^{-2}, 10^{-1}, 1, 5, 10, 15, 20, 25, 30, 35, 100, 10^{3}, 10^{4}, 10^{5}, 10^{6}]
\end{equation*}
and equal $\eta$ weights. We use learning rate between $[0.01, 0.015]$ and $\gamma \in [0.1, 0.3]$. 
For DANN we use learning rate between $[0.01, 0.015]$ and $\gamma \in [0.15, 0.25]$.
For DSN w/ DANN and DSN w/ MMD we use a constant initial learning rate of $0.01$  use the hyperparameters in the range of: $\alpha \in [0.01, 0.15], \beta \in [0.05, 0.075], \gamma \in [0.25, 0.3]$, whereas for DNS w/ CorReg we use $\gamma \in [20, 100]$. For the GTSRB experiment we use $\alpha \in [0.01, 0.015]$. In all cases we use an exponential decay of $0.95$ on the learning rate every $20,000$ iterations. For the LINEMOD experiments we use $\xi = 0.125$.

% \begin{table}[h]
% \centering
% \caption{The parameters that are not universal across experiments and not mentioned in the main paper.}
% \label{tab:parameters}
% \begin{tabular}{ | l | l | l | l | l | }
% \hline
% Hyperparameter   & MNIST to & Synth Digits to & SVHN to  & Synth Signs to \\
%  & MNIST-M  & SVHN        & MNIST & GTSRB     \\ \hline \hline
% Source-only &&&&\\ \hline \hline
% CORAL &&&&\\ \hline
% CorReg&&&&     \\ \hline\hline
% MMD   & 0.01,,,0.253 &&& \\ \hline 
% DANN &&&& \\ \hline
% DSN w/ CorReg &&&& \\ \hline
% DSN w/ MMD&&&& \\ \hline
% DSN w/ DANN &  0.01, 0.0135, 0.05, 0.273 &&& \\ \hline\hline
% Target-only  & &&&  \\ \hline
% \end{tabular}
% \end{table}

\clearpage

\end{document}